\documentclass[final]{cvpr}
\usepackage{nopageno}

\usepackage{times}
\usepackage{epsfig}
\usepackage{graphicx}
\usepackage{amsmath}
\usepackage{amssymb}
\usepackage{subfig}

\usepackage{enumitem}
\usepackage{booktabs}
\usepackage{bm}
\usepackage{wrapfig}
\usepackage[font=small,labelfont=small]{caption}

\usepackage{booktabs}
\usepackage{graphicx}
\usepackage[table,xcdraw,dvipsnames]{xcolor}
\usepackage{array}

\newcolumntype{x}[1]{>{\centering\let\newline\\\arraybackslash\hspace{0pt}}p{#1}}

\usepackage{algorithm}
\usepackage[noend]{algpseudocode}

\allowdisplaybreaks

\newcommand{\cls}[1]{{\small\texttt{#1}}}

\newcommand{\customsubsection}[1]{%
  \par
  \pagebreak[2]%
  \refstepcounter{subsection}%
    \everypar={%
      {\setbox0=\lastbox}
      \addcontentsline{toc}{subsection}{%
        {\protect\makebox[0.3in][r]{\thesubsubsection.} \hspace*{3pt}#1}}%
      \textbf{\thesubsection\space\space{#1}\space}%
      \everypar={}%
    }%
  \ignorespaces
}

\usepackage[pagebackref=true,breaklinks=true,colorlinks,bookmarks=false]{hyperref}

\def\cvprPaperID{8142} 
\def\confYear{CVPR 2021}

\newcommand{\nC}[0]{\textrm{C}}
\newcommand{\nQ}[0]{\textrm{Q}}

\definecolor{iODBlue}{RGB}{220, 232, 250}

\begin{document}

\title{Towards \OWOD \vspace{-10pt}}

\author{K J Joseph$^{\dagger \ddag}$,  Salman Khan$^{\ddag \star}$, Fahad Shahbaz Khan$^{\ddag \diamond}$, Vineeth N Balasubramanian$^{\dagger}$\\
\normalsize$^{\dagger}$Indian Institute of Technology Hyderabad, India \quad $^{\ddag}$Mohamed bin Zayed University of AI, UAE\\
\normalsize $^{\star}$Australian National University, Australia \quad $^{\diamond}$Linköping University, Sweden\\
{\tt\small \{cs17m18p100001,vineethnb\}@iith.ac.in, \{salman.khan,fahad.khan\}@mbzuai.ac.ae}
\vspace{-10pt}
}

\newcommand{\OWOD}{Open World Object Detection\xspace}

\newcommand{\owod}{open world object detection\xspace}

\newcommand{\OSOD}{Open Set Object Detection\xspace}

\newcommand{\osod}{open set object detection\xspace}

\newcommand{\os}{open set\xspace}

\newcommand{\ow}{open world\xspace}

\newcommand{\OS}{Open Set\xspace}

\newcommand{\OW}{Open World\xspace}

\newcommand{\method}{ORE\xspace}

\maketitle

\begin{abstract}\vspace{-0.5em}
   Humans have a natural instinct to identify unknown object instances in their environments. 
   The intrinsic curiosity about these unknown instances aids in learning about them, when the corresponding knowledge is eventually available. This motivates us to propose a novel computer vision problem called: `Open World Object Detection', where a model is tasked to: 1) identify objects that have not been introduced to it as `unknown', without explicit supervision to do so, and 2) incrementally learn these identified unknown categories without forgetting previously learned classes, when the corresponding labels are progressively received. We formulate the problem, introduce a strong evaluation protocol and provide a novel solution, which we call \method: \underline{O}pen Wo\underline{r}ld Object D\underline{e}tector, based on contrastive clustering and energy based unknown identification. Our experimental evaluation and ablation studies analyse the efficacy of \method in achieving \OW objectives. As an interesting by-product, we find that identifying and characterising unknown instances helps to reduce confusion in an incremental object detection setting, where we achieve state-of-the-art performance, with no extra methodological effort. We hope that our work will attract further research into this newly
   identified, yet crucial research direction.\footnote{Source code: \footnotesize\url{https://github.com/JosephKJ/OWOD}} 
\end{abstract}

\vspace{-1em}
\section{Introduction}\label{sec:introduction}
Deep learning has accelerated progress in Object Detection research \cite{girshick2015fast,ren2015faster,he2017mask,lin2017focal,redmon2016you}, where a model is tasked to identify and localise objects in an image. 
All existing approaches work under a strong assumption that all the classes that are to be detected would be available at training phase. Two challenging scenarios arises when we relax this assumption:
1) A test image might contain objects from unknown classes, which should be classified as \emph{unknown}. 
2) As and when information (labels) about such identified unknowns become available, the model should be able to incrementally learn the new class. 
Research in developmental psychology \cite{meacham1983wisdom,livio2017makes} finds out that the ability to identify what one doesn't know, is key in captivating curiosity. Such a curiosity fuels the desire to learn new things \cite{engel2011children,grazer2016curious}. 
This motivates us to propose a new problem where a model should be able to identify instances of unknown objects as unknown and subsequently learns to recognise them when training data progressively arrives, in a \textit{unified} way. We call this problem setting as \textit{\OWOD}.  

\begin{figure}
\includegraphics[width=\columnwidth]{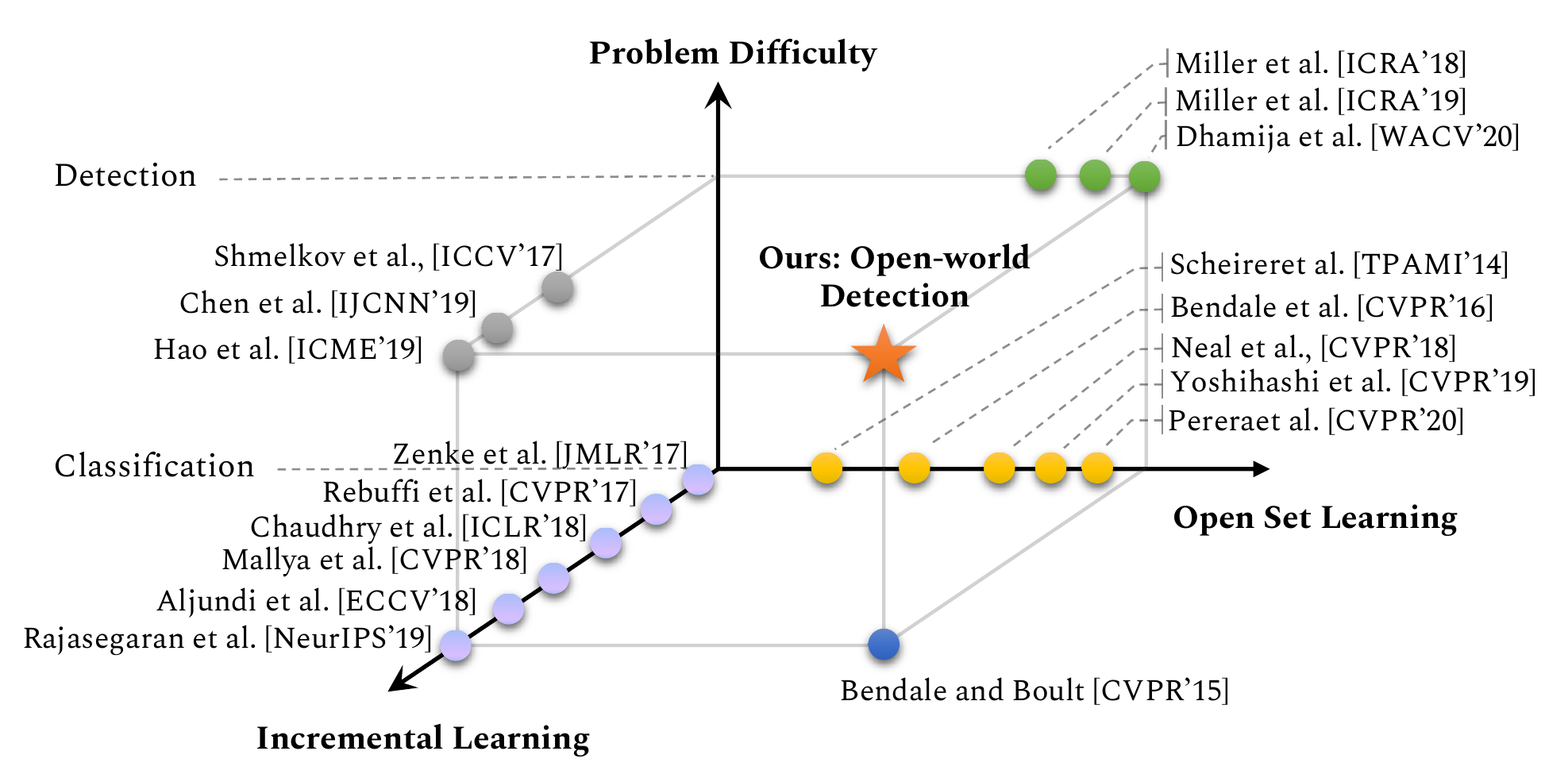}
\caption{\OWOD ({\color{Orange}$\bigstar$}) is a novel problem that has not been formally defined and addressed so far. Though related to the \OS and \OW classification, \OWOD offers its own unique challenges, which when addressed, improves the practicality of object detectors.}
\label{fig:related_works}
\end{figure}

The number of classes that are annotated in standard vision datasets like Pascal VOC \cite{everingham2010pascal} and MS-COCO \cite{lin2014microsoft} are very low (20 and 80 respectively) when compared to the infinite number of classes that are present in the open world. Recognising an unknown as an unknown requires strong generalization. Scheirer \etal \cite{scheirer2012toward} formalise this as \textit{\OS} classification problem. Henceforth, various methodologies (using 1-vs-rest SVMs and deep learning models) has been formulated to address this challenging setting. 
Bendale \etal \cite{bendale2015towards} extend \OS to an \textit{\OW} classification setting by additionally updating the image classifier to recognise the identified new unknown classes. 
Interestingly, as seen in Fig.~\ref{fig:related_works}, Open World object detection is unexplored, owing to the difficulty of the problem setting.


The advances in \OS and \OW image classification cannot be trivially adapted to \OS and \OW object detection, because of a fundamental difference in the problem setting: 
\textit{The object detector is trained to detect unknown objects as background.} Instances of many unknown classes would have been already introduced to the object detector along with known objects. As they are not labelled, these unknown instances would be explicitly learned as background, while training the detection model.
Dhamija \etal \cite{dhamija2020overlooked} find that even with this extra training signal, the state-of-the-art object detectors result in false positive detections, where the unknown objects end up being classified as one of the known classes, often with very high probability. Miller \etal \cite{miller2018dropout} propose to use dropout sampling to get an estimate of the uncertainty of the object detection prediction. This is the only peer-reviewed research work in the \os object detection literature.
Our proposed \OWOD goes a step further to incrementally learn the new classes, once they are detected as unknown and an oracle provides labels for the objects of interest among all the unknowns. To the best of our knowledge this has not been tried in the literature. 

The \OWOD setting is much more natural than the existing closed-world, static-learning setting. The world is diverse and dynamic in the number, type and configurations of novel classes. It would be naive to assume that all the classes to expect at inference are seen during training. Practical deployments of detection systems in robotics, self-driving cars, plant phenotyping, healthcare and surveillance cannot afford to have complete knowledge on what classes to expect at inference time, while being trained in-house. The most natural and realistic behavior that one can expect from an object detection algorithm deployed in such settings would be to confidently predict an unknown object as unknown, and known objects into the corresponding classes. As and when more information about the identified unknown classes becomes available, the system should be able to incorporate them into its existing knowledge base. This would define a smart object detection system, and ours is an effort towards achieving this goal.

\noindent The key contributions of our work are:
\begin{itemize}[leftmargin=*,topsep=0pt, noitemsep]
    \item We introduce a novel problem setting, \OWOD, which models the real-world more closely.
    \item We develop a novel methodology, called \method, based on contrastive clustering, an unknown-aware proposal network and energy based unknown identification to address the challenges of \ow detection.
    \item We introduce a comprehensive experimental setting, which helps to measure the \ow characteristics of an object detector, and benchmark \method on it against competitive baseline methods.
    \item As an interesting by-product, the proposed methodology achieves state-of-the-art performance on Incremental Object Detection, even though not primarily designed for it. 
\end{itemize}

\section{Related Work}\label{sec:related_works}
\noindent\textbf{\OS Classification:} The \os setting considers knowledge acquired through training set to be incomplete, thus new unknown classes can be encountered during testing.
Scheirer \etal \cite{scheirer2013toward}  developed open set classifiers in a one-vs-rest setting to balance the performance and the risk of labeling a sample far from the known training examples (termed as open space risk). Follow up works \cite{jain2014multi, scheirer2014probability} extended the open set framework to multi-class classifier setting with probabilistic models to account for the fading away classifier confidences in case of unknown classes.

Bendale and Boult \cite{bendale2016towards} identified unknowns in the feature space of deep networks and used a Weibull distribution to estimate the set risk (called OpenMax classifier).  A generative version of OpenMax was proposed in \cite{ge2017generative} by synthesizing novel class images. Liu \etal \cite{liu2019large} considered a long-tailed recognition setting where majority, minority and unknown classes coexist. They developed a metric learning framework 
identify unseen classes as unknown. In similar spirit, several dedicated approaches target on detecting the out of distribution samples \cite{liang2018enhancing} or novelties \cite{pidhorskyi2018generative}. Recently, self-supervised learning \cite{Perera_2020_CVPR} and unsupervised learning with reconstruction \cite{Yoshihashi_2019_CVPR} have been explored for open set recognition. However, while these works can recognize unknown instances, they cannot dynamically update themselves in an incremental fashion over multiple training episodes.  Further, our energy based unknown detection approach has not been explored before.

\noindent\textbf{\OW Classification:} \cite{bendale2015towards} first proposed the open world setting for image recognition. Instead of a static classifier trained on a fixed set of classes, they proposed a more flexible setting where knowns and unknowns both coexist. The model can recognize both types of objects and adaptively improve itself when new labels for unknown are provided. Their approach extends Nearest Class Mean classifier to operate in an open world setting by re-calibrating the class probabilities to balance open space risk. \cite{pernici2018memory} studies open world face identity learning while \cite{xu2019open} proposed to use an exemplar set of seen classes to match them against a new sample, and rejects it in case of a low match with all previously known classes. 
However, they don’t test on image classification benchmarks and study product classification in e-commerce applications.

\noindent\textbf{\OS Detection:} Dhamija \etal \cite{dhamija2020overlooked} formally studied the impact of open set setting on popular object detectors. They noticed that the state of the art object detectors often classify unknown classes with high confidence to seen classes. This is despite the fact that the detectors are explicitly trained with a background class \cite{ren2016faster, girshick2015fast, liu2016ssd} and/or apply one-vs-rest classifiers to model each  class \cite{girshick2014rich, lin2017focal}. A dedicated body of work \cite{miller2018dropout, miller2019evaluating, hall2020probabilistic} focuses on developing measures of (spatial and semantic) uncertainty in object detectors to reject unknown classes. E.g., \cite{miller2018dropout,miller2019evaluating}  uses Monte Carlo Dropout \cite{gal2016dropout} sampling in a SSD detector to obtain uncertainty estimates. These methods, however, cannot incrementally adapt their knowledge in a dynamic world.


\section{\OWOD}\label{sec:prob_definition}

\begin{figure*}[t]
\centering
\begin{minipage}[]{0.75\textwidth}
\includegraphics[width=1\linewidth]{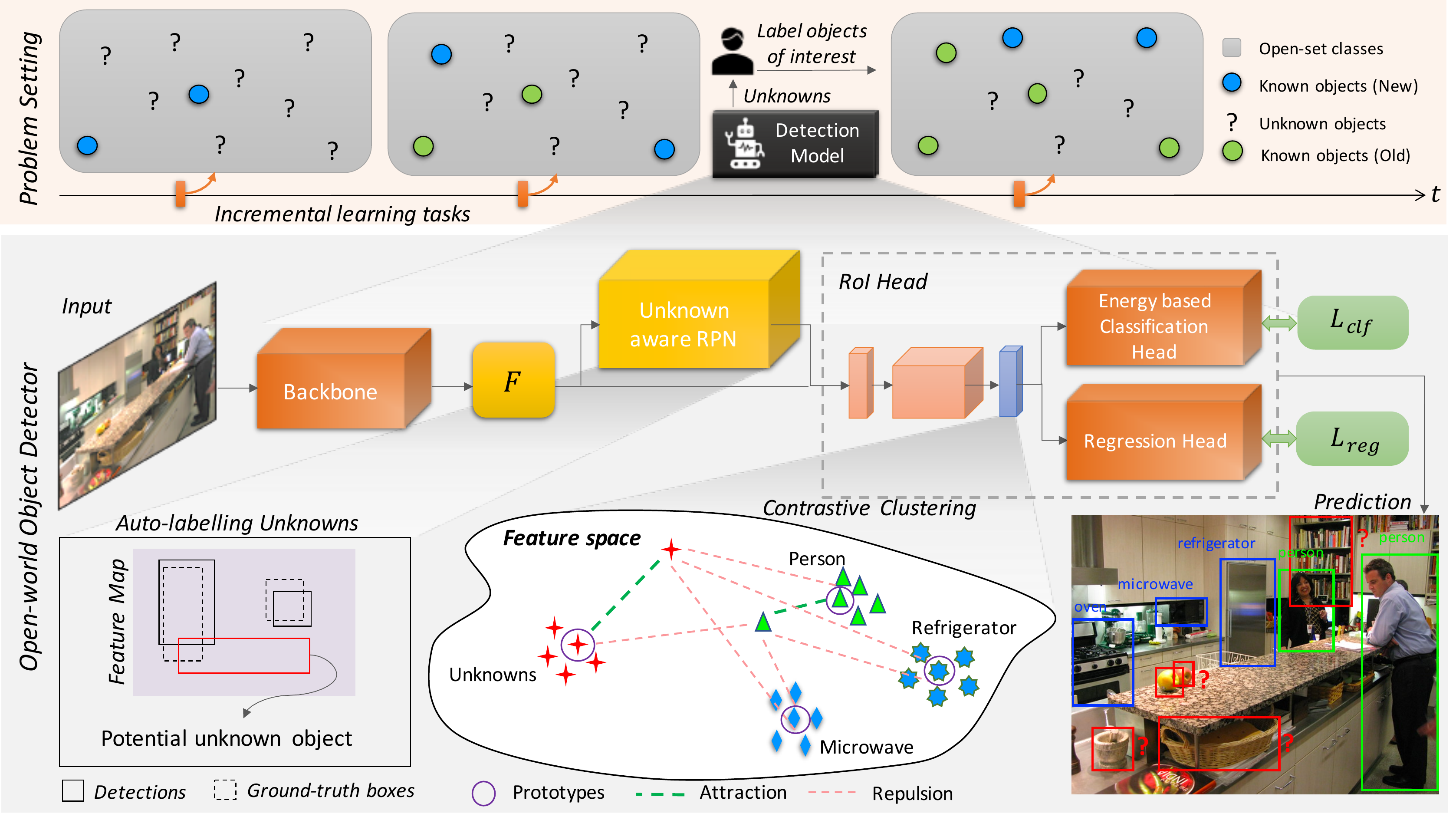}
\end{minipage}
\begin{minipage}[]{0.24\textwidth}
\captionof{figure}{\emph{Approach Overview:} \emph{Top row:} At each incremental learning step, the model identifies unknown objects (denoted by `?'), which are progressively labelled (as blue circles) and added to the existing knowledge base (green circles). \emph{Bottom row:} Our open world object detection model identifies potential unknown objects using an energy-based classification head and the unknown-aware RPN. Further, we perform contrastive learning in the feature space to learn discriminative clusters and can flexibly add new classes in a continual manner without forgetting the previous classes. 
}
\label{fig:pipeline}
\end{minipage}
\vspace{-5pt}
\end{figure*}


Let us formalise the definition of \OWOD in this section.
At any time $t$, we consider the set of known object classes as $\mathcal{K}^t = \{1, 2, .. , \nC\} \subset \mathbb{N}^+$ where $\mathbb{N}^+$ denotes the set of positive integers. In order to realistically model the dynamics of real world, we also assume that their exists a set of unknown classes  $\mathcal{U} = \{\nC + 1, ...  \}$, which may be  encountered during inference. The known object classes $\mathcal{K}_t$ are assumed to be labeled in the dataset  $\mathcal{D}^t = \{ \mathbf{X}^t, \mathbf{Y}^t\}$ where $\mathbf{X}$ and $\mathbf{Y}$ denote the input images and labels respectively. The input image set comprises of $M$ training images, $\mathbf{X}^t = \{\bm{I}_1, \ldots, \bm{I}_{M} \}$ and associated object labels for each image forms the label set $\mathbf{Y}^t = \{\bm{Y}_1, \ldots, \bm{Y}_{M} \}$. Each $\bm{Y}_i = \{\bm{y}_1, \bm{y}_2, .., \bm{y}_K \} $ encodes a set of $K$ object instances with their class labels and locations i.e., $\bm{y}_k = [l_k, x_k, y_k, w_k, h_k]$, where $l_k \in \mathcal{K}^t$ and $x_k, y_k, w_k, h_k$ denote the bounding box center coordinates, width and height respectively.
 
The \textit{\OWOD} setting considers an object detection model $\mathcal{M}_{\nC}$ that is trained to detect all the previously encountered $\nC$ object classes. Importantly, the model $\mathcal{M}_{\mathrm{C}}$ is able to identify a test instance belonging to any of the known $\mathrm{C}$ classes, and can also recognize a new or unseen class instance by classifying it as an \emph{unknown}, 
denoted by a label zero (0). The unknown set of instances $\mathbf{U}^t$ can then be forwarded to a human user who can identify $n$ new classes of interest (among a potentially large number of unknowns) and provide their training examples. The learner incrementally adds $n$ new classes and updates itself to produce an updated model $\mathcal{M}_{\nC+n}$ without retraining from scratch on the whole dataset. The known class set is also updated $\mathcal{K}_{t+1} = \mathcal{K}_t  + \{\nC + 1, \ldots, \nC + n\} $. This cycle continues over the life of the object detector, where it adaptively updates itself  with new knowledge. 
The problem setting is illustrated in the top row of Fig.~\ref{fig:pipeline}.

\section{\method: \underline{O}pen Wo\underline{r}ld Object D\underline{e}tector}\label{sec:ore_methodology}
A successful approach for \OWOD should be able to identify unknown instances without explicit supervision and defy forgetting of earlier instances  when labels of these identified novel instances are presented to the model for knowledge upgradation (without retraining from scratch). We propose a solution, \method 
which addresses both these challenges in a unified manner. 

Neural networks are universal function approximators \cite{hornik1989multilayer}, which learn a mapping between an input and the output through a series of hidden layers. The latent representation learned in these hidden layers directly controls how each function is realised. We hypothesise that learning clear discrimination between classes in the latent space of object detectors could have two fold effect. \emph{First}, it helps the model to identify how the feature representation of an unknown instance is different from the other known instances, which helps identify an unknown instance as a novelty. \emph{Second}, it facilitates learning feature representations for the new class instances without overlapping with the previous classes in the latent space, which helps towards incrementally learning without forgetting. 
The key component that helps us realise this is our proposed \textit{contrastive clustering} in the latent space, which we elaborate in Sec. \ref{sec:contrastive_clustering}.

To optimally cluster the unknowns using contrastive clustering, we need to have supervision on what an unknown instance is. It is infeasible to manually annotate even a small subset of the potentially infinite set of unknown classes. To counter this, we propose an auto-labelling mechanism based on the Region Proposal Network \cite{ren2015faster} to pseudo-label unknown instances, as explained in Sec. \ref{sec:autolabelling_unknown}. 
The inherent separation of auto-labelled unknown instances in the latent space helps our energy based classification head to differentiate between the known and unknown instances. As elucidated in Sec. \ref{sec:energy_based_unk_identification}, we find that Helmholtz free energy is higher for unknown instances.

Fig.~\ref{fig:pipeline} shows the high-level architectural overview of \method. We choose Faster R-CNN \cite{ren2015faster} as the base detector as Dhamija \etal.  \cite{dhamija2020overlooked} has found that it has better \os performance when compared against one-stage Retina-Net detector \cite{lin2017focal} and objectness based YOLO  detector \cite{redmon2016you}. 
Faster R-CNN \cite{ren2015faster} is a two stage object detector. In the first stage, a class-agnostic Region Proposal Network (RPN) proposes potential regions which might have an object from the feature maps coming from a shared backbone network. The second stage classifies and adjusts the bounding box coordinates of each of the proposed region. 
The features that are generated by the residual block in the Region of Interest (RoI) head are contrastively clustered.
The RPN and the classification head is adapted to auto-label and identify unknowns respectively. We explain each of these coherent constituent components, in the following subsections: 

\subsection{Contrastive Clustering} \label{sec:contrastive_clustering}
Class separation in the latent space would be an ideal characteristic for an \OW methodology to identify unknowns. A natural way to enforce this would be to model it as a contrastive clustering problem, where instances of same class would be forced to remain close-by, while instances of dissimilar class would be pushed far apart. 

For each known class $i \in \mathcal{K}^t$, we maintain a prototype vector $\bm{p}_i$. Let $\bm{f}_c \in \mathbb{R}^d$ be a feature vector that is generated by an intermediate layer of the object detector, for an object of class $c$. We define the contrastive loss as follows:
\begin{align}\label{eqn:clustering_loss}
    \mathcal{L}_{cont}(\bm{f}_c) & = \sum_{i = 0}^{\nC} \ell(\bm{f}_c, \bm{p}_i), \text{ where,} \\
    \ell(\bm{f}_c, \bm{p}_i) & = \begin{cases}      \mathcal{D}(\bm{f}_c, \bm{p}_i) & i = c\\       \max\{0, \Delta - \mathcal{D}(\bm{f}_c, \bm{p}_i)\} & \text{otherwise}   \end{cases} \notag
\end{align}
where $\mathcal{D}$ is any distance function and $\Delta$ defines how close a similar and dissimilar item can be. Minimizing this loss would ensure the desired class separation in the latent space. 

Mean of feature vectors corresponding to each class is used to create the set of class prototypes: $\mathcal{P} = \{\bm{p}_0 \cdots \bm{p}_\nC\}$. Maintaining each prototype vector is a crucial component of \method. As the whole network is trained end-to-end, the class prototypes should also gradually evolve, as the constituent features change gradually (as stochastic gradient descent  updates weights by a small step in each iteration). We maintain a fixed-length queue $\bm{q}_i$, per class for storing the corresponding features. A feature store $\mathcal{F}_{store} = \{\bm{q}_0 \cdots \bm{q}_\nC\}$, stores the class specific features in the corresponding queues. This is a scalable approach for keeping track of how the feature vectors evolve with training, as the number of feature vectors that are stored is bounded by $\nC \times \nQ$, where $\nQ$ is the maximum size of the queue.

Algorithm \ref{algo:get_clustering_loss} provides an overview on how class prototypes are managed while computing the clustering loss. We start computing the loss only after a certain number of burn-in iterations ($I_b$) are completed. This allows the initial feature embeddings to mature themselves to encode class information. Since then, we compute the clustering loss using Eqn.~\ref{eqn:clustering_loss}. After every $ I_p$ iterations, a set of new class prototypes $\mathcal{P}_{new}$ is computed (line 8). Then the existing prototypes $\mathcal{P}$ are updated by weighing $\mathcal{P}$ and $\mathcal{P}_{new}$ with a momentum parameter $\eta$. This allows the class prototypes to evolve gradually keeping track of previous context. The computed clustering loss is added to the standard detection loss and back-propagated to learn the network end-to-end.

\begin{algorithm}\small
\caption{Algorithm \textsc{ComputeClusteringLoss}}
\label{algo:get_clustering_loss}
\begin{algorithmic}[1]
\Require{Input feature for which loss is computed: $\bm{f}_c$; Feature store: $\mathcal{F}_{store}$; Current iteration: $i$; Class prototypes: $\mathcal{P} = \{\bm{p}_0 \cdots \bm{p}_\nC\}$; Momentum parameter: $\eta$.
}
\State Initialise $\mathcal{P}$ if it is the first iteration.
\State $\mathcal{L}_{cont}$ $\leftarrow$ 0
\If{$i== I_b$}
\State $\mathcal{P} \leftarrow$ class-wise mean of items in $\mathcal{F}_{Store}$. 
\State $\mathcal{L}_{cont}$ $\leftarrow$ Compute using $\bm{f}_c$, $\mathcal{P}$ and Eqn. \ref{eqn:clustering_loss}.  
\ElsIf{$i> $  $I_b$}
 \If{$i\% I_p == 0$}
 \State $\mathcal{P}_{new} \leftarrow$ class-wise mean of  items in $\mathcal{F}_{Store}$. 
 \State $\mathcal{P} \leftarrow \eta \mathcal{P} + (1-\eta)\mathcal{P}_{new}$
 \EndIf
 \State $\mathcal{L}_{cont}$ $\leftarrow$ Compute using $\bm{f}_c$, $\mathcal{P}$ and Eqn. \ref{eqn:clustering_loss}.
\EndIf
\State \Return $\mathcal{L}_{cont}$
\end{algorithmic}
\end{algorithm}

\subsection{Auto-labelling Unknowns with RPN}\label{sec:autolabelling_unknown}
While computing the clustering loss with Eqn.~\ref{eqn:clustering_loss}, we contrast the input feature vector $\bm{f}_c$ against prototype vectors, which include a prototype for unknown objects too ($c\in \{0,1,..,\nC\}$  where $0$ refers to the unknown class). This would require unknown object instances to be labelled with \cls{unknown} ground truth class, which is not practically feasible owing to the arduous task of re-annotating \underline{all} instances of each image in already annotated large-scale datasets.

As a surrogate, we propose to automatically label some of the objects in the image as a potential unknown object. For this, we rely on the fact that Region Proposal Network (RPN) is class agnostic.
Given an input image, the RPN generates a set of bounding box predictions for foreground and background instances, along with the corresponding objectness scores.
We label those proposals that have high objectness score, but do not overlap with a ground-truth object as a potential unknown object. Simply put, we select the top-k background region proposals, sorted by its objectness scores, as unknown objects. 
This seemingly simple heuristic achieves good performance as demonstrated in Sec.~\ref{sec:expr_and_results}.

\subsection{Energy Based Unknown Identifier}\label{sec:energy_based_unk_identification}

Given the features ($\bm{f} \in F$) in the latent space $F$ and their corresponding labels $l \in L$, we seek to learn an energy function $E(F,L)$. Our formulation is based on the Energy based models (EBMs) \cite{lecun2006tutorial} that learn a function $E(\cdot)$ to estimates the compatibility between observed variables $F$ and possible set of output variables $L$ using a single output scalar i.e., $E(\bm{f}): \mathbb{R}^d \rightarrow \mathbb{R}$. 
The intrinsic capability of EBMs to assign low energy values to in-distribution data and vice-versa motivates us to use an energy measure to characterize whether a sample is from an unknown class.

\begin{figure}[t]
\centering
\includegraphics[width=0.9\linewidth]{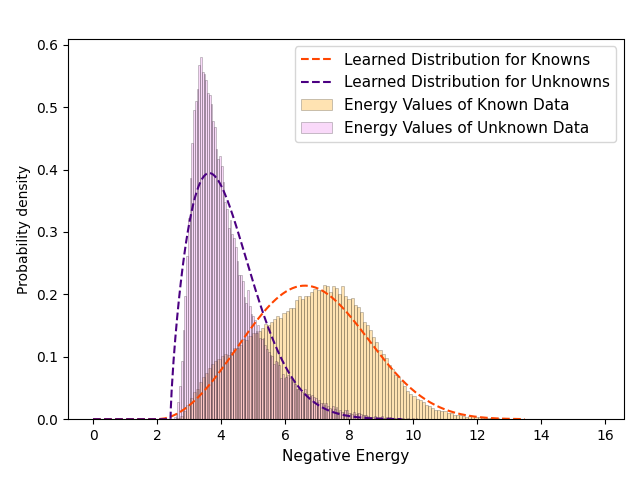}
\vspace{-13pt}
\caption{\small
The energy values of the known and unknown data-points exhibit clear separation as seen above. We fit a Weibull distribution on each of them and use these for identifying unseen known and unknown samples, as explained in Sec.~\ref{sec:energy_based_unk_identification}.}
\label{fig:energy_plots}
\end{figure}

Specifically, we use the Helmholtz free energy formulation where energies for all values in $L$ are combined,
\begin{equation}
    E(\bm{f}) = -T \log \int_{l'} \exp\bigg({-\frac{E(\bm{f}, l')}{T}} \bigg),
    \label{eqn:free_energy}
\end{equation}
where $T$ is the temperature parameter. There exists a simple relation between the network outputs after the softmax layer and the Gibbs distribution of class specific energy values \cite{liu2020energy}. This can be formulated as, 
\begin{equation}
    p(l | \bm{f}) = \frac{ \exp(\frac{g_{l}(\bm{f})}{T}) }{\sum_{i=1}^{\nC} 
    \exp(\frac{g_{i}(\bm{f})}{T})} =
    \frac{ \exp(-\frac{E(\bm{f}, l)}{T})}{ \exp (- \frac{E(\bm{f})}{T})}
\end{equation}
where $p(l | \bm{f})$ is the probability density for a label $l$, $g_l(\bm{f})$ is the $l^{th}$ classification logit of the classification head $g(.)$. Using this correspondence, we define free energy of our classification models in terms of their logits as follows:
\vspace{-5pt}
\begin{equation}
    E(\bm{f}; g) = -T \log \sum_{i = 1}^{\nC} \exp(\frac{g_i(\bm{f})}{T}).
    \vspace{-5pt}
    \label{eqn:energy}
\end{equation}
The above equation provides us a natural way to transform the classification head of the standard Faster R-CNN~\cite{ren2015faster} to an energy function. Due to the clear separation that we enforce in the latent space with the contrastive clustering, we see a clear separation in the energy level of the known class data-points and unknown data-points as illustrated in Fig. \ref{fig:energy_plots}. In light of this trend, we model the energy distribution of the known and unknown energy values  $\xi_{kn}(\bm{f})$ and $\xi_{unk}(\bm{f})$, with a set of shifted Weibull distributions. These distributions were found to fit the energy data of a small held out validation set (with both knowns and unknowns instances) very well, when compared to Gamma, Exponential and Normal distributions. The learned distributions can be used to label a prediction as unknown if $\xi_{kn}(\bm{f}) < \xi_{unk}(\bm{f})$.

\subsection{Alleviating Forgetting}
After the identification of unknowns, an important requisite for an \ow  detector is to be able to learn new classes, when the labeled examples of some of the unknown classes of interest are provided. 
Importantly, the training data for the previous tasks will not be present at this stage since retraining from scratch is not a feasible solution. Training with only the new class instances will lead to catastrophic forgetting \cite{mccloskey1989catastrophic,french1999catastrophic} of the previous classes.
We note that a number of involved approaches have been developed to alleviate such forgetting, including methods based on parameter regularization \cite{aljundi2018memory,kirkpatrick2017overcoming,li2018learning,zenke2017continual}, exemplar replay \cite{AGEM,rebuffi2017icarl,lopez2017gradient,castro2018end}, dynamically expanding networks \cite{mallya2018packnet,serra2018overcoming,rusu2016progressive} and meta-learning \cite{rajasegaran2020itaml,kj2020meta}. 

We build on the recent insights from \cite{prabhu2020gdumb,knoblauch2020optimal,wang2020frustratingly} which compare the importance of example replay against other more complex solutions. Specifically, Prabhu \etal \cite{prabhu2020gdumb} retrospects the progress made by the complex continual learning methodologies and show that a greedy exemplar selection strategy for replay in incremental learning  consistently outperforms the state-of-the-art methods by a large margin. Knoblauch \etal \cite{knoblauch2020optimal} develops a theoretical justification for the unwarranted power of replay methods. They prove that an optimal continual learner solves an NP-hard problem and requires infinite memory. The effectiveness of storing few examples and replaying has been found effective in the related few-shot object detection setting by Wang \etal \cite{wang2020frustratingly}. These motivates us to use a relatively simple methodology for \method to mitigate forgetting i.e., we store a balanced set of exemplars and finetune the model after each incremental step on these. At each point, we ensure that a minimum of $N_{ex}$ instances for each class are present in the exemplar set.


\section{Experiments and Results} \label{sec:expr_and_results}
We propose a comprehensive evaluation protocol to study the performance of an open world detector to identify unknowns, detect known classes and progressively learn new classes when labels are provided for some unknowns. 

\subsection{\OW Evaluation Protocol} \label{sec:evaluation_protocol}
\noindent \textbf{Data split:} We group classes into a set of tasks $\mathcal{T} = \{T_1, \cdots T_t, \cdots\}$. All the classes of a specific task will be introduced to the system at a point of time $t$. While learning $T_t$, all the classes of $\{T_\tau:\tau {<} t\}$ will be treated as known and  $\{T_\tau:\tau {>} t\}$ would be treated as unknown. For a concrete instantiation of this protocol, we consider classes from Pascal VOC \cite{everingham2010pascal} and MS-COCO \cite{lin2014microsoft}. We group all VOC classes and data as the first task $T_1$. The remaining $60$ classes of MS-COCO \cite{lin2014microsoft} are grouped into three successive tasks with semantic drifts (see Tab. \ref{tab:data_split}). 
All images  which correspond to the above split from  Pascal VOC and MS-COCO train-sets form the training data. 
For evaluation, we use the Pascal VOC test split and MS-COCO val split. $1$k images from training data of each task is kept aside for validation.
Data splits and codes can be found at {\small\url{https://github.com/JosephKJ/OWOD}}.

\noindent \textbf{Evaluation metrics:} 
Since an unknown object easily gets confused as a known object, we use the Wilderness Impact (WI) metric \cite{dhamija2020overlooked} to explicitly characterises this behaviour.
\begin{equation}
    \text{Wilderness Impact} \,  (WI) = \frac{P_{\mathcal{K}}}{P_{\mathcal{K} \cup \mathcal{U}}} - 1,
\end{equation}
where $P_{\mathcal{K}}$ refers to the precision of the model when evaluated on known classes and $P_{\mathcal{K} \cup \mathcal{U}}$ is the precision when evaluated on known and unknown classes, measured at a recall level $R$ (0.8 in all experiments). Ideally, WI should be less as the precision must not drop when unknown objects are added to the test set. Besides WI, we also use  Absolute Open-Set Error (A-OSE) \cite{miller2018dropout}
to report the number count of unknown objects that get wrongly classified as any of the known class. 
Both WI and A-OSE implicitly measure how effective the model is in handling unknown objects. 

In order to quantify incremental learning capability of the model in the presence of new labeled classes, we measure the mean Average Precision (mAP) at IoU threshold of 0.5 (consistent with the existing literature \cite{shmelkov2017incremental,PENG2020109}).  

\begin{table}
\centering
\resizebox{0.48\textwidth}{!}{%
\begin{tabular}{@{}l|cccc@{}}
\toprule
 & Task 1 & Task 2 & Task 3 & Task 4 \\ \midrule
Semantic split & \begin{tabular}[c]{@{}c@{}}VOC \\ Classes\end{tabular} & \begin{tabular}[c]{@{}c@{}}Outdoor, Accessories, \\ Appliance, Truck\end{tabular} & \begin{tabular}[c]{@{}c@{}}Sports, \\ Food\end{tabular} & \begin{tabular}[c]{@{}c@{}}Electronic, Indoor, \\ Kitchen, Furniture\end{tabular} \\\midrule
\# training images & 16551 & 45520 & 39402 & 40260 \\
\# test images & 4952 & 1914 & 1642 & 1738 \\
\# train instances & 47223 & 113741 & 114452 & 138996 \\
\# test instances & 14976 & 4966 & 4826 & 6039 \\ \bottomrule
\end{tabular}%
}\vspace{-0.5em}
\caption{The table shows task composition in the proposed \OW evaluation protocol. The semantics of each task and the number of images and instances (objects) across splits are shown.}
\label{tab:data_split}
\end{table}

\begin{table*}[!htp]\setlength{\tabcolsep}{2pt}
\resizebox{\textwidth}{!}{%
\begin{tabular}{@{}l|c|c|c|c|c|ccc|c|c|ccc|ccc@{}}
\toprule
 Task IDs ($\rightarrow$)& \multicolumn{3}{c|}{Task 1} & \multicolumn{5}{c|}{Task 2} & \multicolumn{5}{c|}{Task 3} & \multicolumn{3}{c}{Task 4} \\ \midrule
 & \cellcolor[HTML]{F3F3F3}WI & \cellcolor[HTML]{F3F3F3}A-OSE  & \multicolumn{1}{c|}{mAP ($\uparrow$)} & \cellcolor[HTML]{F3F3F3}WI & \cellcolor[HTML]{F3F3F3}A-OSE  & \multicolumn{3}{c|}{mAP ($\uparrow$)} & \cellcolor[HTML]{F3F3F3}WI  & \cellcolor[HTML]{F3F3F3}A-OSE & \multicolumn{3}{c|}{mAP ($\uparrow$)} & \multicolumn{3}{c}{mAP ($\uparrow$)} \\ \cmidrule(lr){4-4} \cmidrule(lr){7-9}  \cmidrule(lr){12-14}\cmidrule(lr){15-17}
 & \cellcolor[HTML]{F3F3F3}($\downarrow$) & \cellcolor[HTML]{F3F3F3}($\downarrow$) & \begin{tabular}[c]{@{}c}Current \\ known\end{tabular} & \cellcolor[HTML]{F3F3F3}($\downarrow$) & \cellcolor[HTML]{F3F3F3}($\downarrow$) & \begin{tabular}[c]{@{}c@{}}Previously\\  known\end{tabular} & \begin{tabular}[c]{@{}c@{}}Current \\ known\end{tabular} & Both & \cellcolor[HTML]{F3F3F3}($\downarrow$) & \cellcolor[HTML]{F3F3F3}($\downarrow$) & \begin{tabular}[c]{@{}c@{}}Previously \\ known\end{tabular} & \begin{tabular}[c]{@{}c@{}}Current \\ known\end{tabular} & Both & \begin{tabular}[c]{@{}c@{}}Previously \\ known\end{tabular} & \begin{tabular}[c]{@{}c@{}}Current \\ known\end{tabular} & Both \\ \midrule
Oracle & \cellcolor[HTML]{F3F3F3}0.02004 & \cellcolor[HTML]{F3F3F3}7080  & 57.76 & \cellcolor[HTML]{F3F3F3}0.0066 & \cellcolor[HTML]{F3F3F3}6717 & 54.99 & 30.31 & 42.65 & \cellcolor[HTML]{F3F3F3}0.0038 & \cellcolor[HTML]{F3F3F3}4237 & 40.23 & 21.51 & 30.87 & 32.52 & 19.27 & 31.71 \\ \midrule
Faster-RCNN & \cellcolor[HTML]{F3F3F3}0.06991 & \cellcolor[HTML]{F3F3F3}13396  & 56.16 & \cellcolor[HTML]{F3F3F3}0.0371 & \cellcolor[HTML]{F3F3F3}12291 & 4.076 & 25.74 & 14.91 & \cellcolor[HTML]{F3F3F3}0.0213 & \cellcolor[HTML]{F3F3F3}9174 & 6.96 & 13.481 & 9.138 & 2.04 & 13.68 & 4.95 \\ \midrule
\begin{tabular}[c]{@{}l@{}}Faster-RCNN\\+ Finetuning\end{tabular} & \multicolumn{3}{c|}{\begin{tabular}[c]{@{}c}Not applicable as incremental\\component is not present in Task 1\end{tabular}} & \cellcolor[HTML]{F3F3F3}0.0375 & \cellcolor[HTML]{F3F3F3}12497 & 51.09 & 23.84 & 37.47 & \cellcolor[HTML]{F3F3F3}0.0279 & \cellcolor[HTML]{F3F3F3}9622 & 35.69 & 11.53 & 27.64 & 29.53 & 12.78 & 25.34 \\ \midrule
ORE & \cellcolor[HTML]{F3F3F3}\textbf{0.02193} & \cellcolor[HTML]{F3F3F3}\textbf{8234}  & \textbf{56.34} & \cellcolor[HTML]{F3F3F3}\textbf{0.0154} & \cellcolor[HTML]{F3F3F3}\textbf{7772} & 52.37 & 25.58 & \textbf{38.98} & \cellcolor[HTML]{F3F3F3}\textbf{0.0081} & \cellcolor[HTML]{F3F3F3}\textbf{6634} & 37.77 & 12.41 & \textbf{29.32} & 30.01 & 13.44 & \textbf{26.66} \\ \bottomrule
\end{tabular}%
}\vspace{-0.5em}
\caption{Here we showcase how \method performs on \OWOD. Wilderness Impact (WI) and Average Open Set Error (A-OSE) quantify how \method handles the unknown classes (\colorbox{ gray!15}{gray} background), whereas Mean Average Precision (mAP) measures how well it detects the known classes (white background). We see that \method consistently outperforms the Faster R-CNN based baseline on all the metrics. Kindly refer to Sec.~\ref{sec:main_results} for more detailed analysis and explanation for the evaluation metrics.
}\vspace{-0.8em}
\label{tab:main_table}
\end{table*}

\subsection{Implementation Details} \label{sec:impl_details}

\method re-purposes the standard Faster R-CNN \cite{ren2015faster} object detector with a ResNet-50 \cite{he2016deep} backbone. To handle variable number of classes in the classification head, following incremental classification methods \cite{rajasegaran2020itaml, kj2020meta,AGEM,lopez2017gradient}, we assume a bound on the maximum number of classes to expect, and modify the loss to take into account only the classes of interest. This is done by setting the classification logits of the unseen classes to a large negative value ($v$), thus making their contribution to softmax negligible ($e^{-v} \rightarrow 0$). 

The $2048$-dim feature vector which comes from the last residual block in the RoI Head is used for contrastive clustering. The contrastive loss (defined in Eqn. \ref{eqn:clustering_loss}) is added to the standard Faster R-CNN classification and localization losses and jointly optimised for. 
While learning a task $T_i$, only the classes that are part of $T_i$ will be labelled. While testing $T_i$, all the classes that were previously introduced are labelled along with classes in $T_i$, and  all classes of future tasks will be labelled `\cls{unknown}'. For the exemplar replay, we empirically choose $N_{ex}= 50$. We do a sensitivity analysis on the size of the exemplar memory in Sec.~\ref{sec:discussions_and_analysis}. Further implementation details are provided in supplementary.

\subsection{\OWOD Results}\label{sec:main_results}

Table \ref{tab:main_table} shows how \method compares against Faster R-CNN on the proposed open world evaluation protocol.
An `Oracle' detector has access to all known and unknown labels at any point, and serves as a reference. 
After learning each task, WI and A-OSE metrics are used to quantify how unknown instances are confused with any of the known classes. We see that \method has significantly lower WI and A-OSE scores, owing to an explicit modeling of the unknown.
When unknown classes are progressively labelled in Task 2, we see that the performance of the baseline detector on the known set of classes (quantified via mAP) significantly deteriorates from $56.16  \%$ to $4.076  \%$. The proposed balanced finetuning is able to restore the previous class performance to a respectable level ($51.09\%$) at the cost of increased WI and A-OSE, whereas \method is able to achieve both goals: detect known classes and reduce the effect of unknown comprehensively. Similar trend is seen when Task 3 classes are added. WI and A-OSE scores cannot be measured for Task 4 because of the absence of any unknown ground-truths. We report qualitative results in Fig.~\ref{fig:qual_results} and supplementary section, along with failure case analysis. 
We conduct extensive sensitivity analysis in Sec.~\ref{sec:discussions_and_analysis} and supplementary section.

\subsection{Incremental Object Detection Results}\label{sec:incr_OD_results}

\begin{table*}
\centering\setlength{\tabcolsep}{3pt}
\resizebox{\textwidth}{!}{%
\begin{tabular}{@{}llllllllllllllllllllll@{}}
\toprule
{\color[HTML]{009901} \textbf{10 + 10 setting}} & aero & cycle & bird & boat & bottle & bus & car & cat & chair & cow & table & dog & horse & bike & person & plant & sheep & sofa & train & tv & mAP \\ \midrule
All 20 & 68.5 & 77.2 & 74.2 & 55.6 & 59.7 & 76.5 & 83.1 & 81.5 & 52.1 & 79.8 & \cellcolor[HTML]{DAE8FC}55.1 & \cellcolor[HTML]{DAE8FC}80.9 & \cellcolor[HTML]{DAE8FC}80.1 & \cellcolor[HTML]{DAE8FC}76.8 & \cellcolor[HTML]{DAE8FC}80.5 & \cellcolor[HTML]{DAE8FC}47.1 & \cellcolor[HTML]{DAE8FC}73.1 & \cellcolor[HTML]{DAE8FC}61.2 & \cellcolor[HTML]{DAE8FC}76.9 & \cellcolor[HTML]{DAE8FC}70.3 & 70.51\\
First 10 & 79.3 & 79.7 & 70.2 & 56.4 & 62.4 & 79.6 & 88.6 & 76.6 & 50.1 & 68.9 & \cellcolor[HTML]{DAE8FC}0 & \cellcolor[HTML]{DAE8FC}0 & \cellcolor[HTML]{DAE8FC}0 & \cellcolor[HTML]{DAE8FC}0 & \cellcolor[HTML]{DAE8FC}0 & \cellcolor[HTML]{DAE8FC}0 & \cellcolor[HTML]{DAE8FC}0 & \cellcolor[HTML]{DAE8FC}0 & \cellcolor[HTML]{DAE8FC}0 & \cellcolor[HTML]{DAE8FC}0 & 35.59 \\
New 10 & 7.9 & 0.3 & 5.1 & 3.4 & 0 & 0 & 0.2 & 2.3 & 0.1 & 3.3 & \cellcolor[HTML]{DAE8FC}65 & \cellcolor[HTML]{DAE8FC}69.3 & \cellcolor[HTML]{DAE8FC}81.3 & \cellcolor[HTML]{DAE8FC}76.4 & \cellcolor[HTML]{DAE8FC}83.1 & \cellcolor[HTML]{DAE8FC}47.2 & \cellcolor[HTML]{DAE8FC}67.1 & \cellcolor[HTML]{DAE8FC}68.4 & \cellcolor[HTML]{DAE8FC}76.5 & \cellcolor[HTML]{DAE8FC}69.2 & 36.31 \\ \midrule
ILOD \cite{shmelkov2017incremental} & 69.9 & 70.4 & 69.4 & 54.3 & 48 & 68.7 & 78.9 & 68.4 & 45.5 & 58.1 & \cellcolor[HTML]{DAE8FC}59.7 & \cellcolor[HTML]{DAE8FC}72.7 & \cellcolor[HTML]{DAE8FC}73.5 & \cellcolor[HTML]{DAE8FC}73.2 & \cellcolor[HTML]{DAE8FC}66.3 & \cellcolor[HTML]{DAE8FC}29.5 & \cellcolor[HTML]{DAE8FC}63.4 & \cellcolor[HTML]{DAE8FC}61.6 & \cellcolor[HTML]{DAE8FC}69.3 & \cellcolor[HTML]{DAE8FC}62.2 & 63.15 \\
ILOD + Faster R-CNN & 70.5 & 75.6 & 68.9 & 59.1 & 56.6 & 67.6 & 78.6 & 75.4 & 50.3 & 70.8 & \cellcolor[HTML]{DAE8FC}43.2 & \cellcolor[HTML]{DAE8FC}68.1 & \cellcolor[HTML]{DAE8FC}66.2 & \cellcolor[HTML]{DAE8FC}65.1 & \cellcolor[HTML]{DAE8FC}66.5 & \cellcolor[HTML]{DAE8FC}24.3 & \cellcolor[HTML]{DAE8FC}61.3 & \cellcolor[HTML]{DAE8FC}46.6 & \cellcolor[HTML]{DAE8FC}58.1 & \cellcolor[HTML]{DAE8FC}49.9 & 61.14 \\
Faster ILOD \cite{PENG2020109} & 72.8 & 75.7 & 71.2 & 60.5 & 61.7 & 70.4 & 83.3 & 76.6 & 53.1 & 72.3 & \cellcolor[HTML]{DAE8FC}36.7 & \cellcolor[HTML]{DAE8FC}70.9 & \cellcolor[HTML]{DAE8FC}66.8 & \cellcolor[HTML]{DAE8FC}67.6 & \cellcolor[HTML]{DAE8FC}66.1 & \cellcolor[HTML]{DAE8FC}24.7 & \cellcolor[HTML]{DAE8FC}63.1 & \cellcolor[HTML]{DAE8FC}48.1 & \cellcolor[HTML]{DAE8FC}57.1 & \cellcolor[HTML]{DAE8FC}43.6 & 62.16 \\ \midrule
ORE - (CC  + EBUI) & 53.3 & 69.2 & 62.4 & 51.8 & 52.9 & 73.6 & 83.7 & 71.7 & 42.8 & 66.8 & \cellcolor[HTML]{DAE8FC}46.8 & \cellcolor[HTML]{DAE8FC}59.9 & \cellcolor[HTML]{DAE8FC}65.5 & \cellcolor[HTML]{DAE8FC}66.1 & \cellcolor[HTML]{DAE8FC}68.6 & \cellcolor[HTML]{DAE8FC}29.8 & \cellcolor[HTML]{DAE8FC}55.1 & \cellcolor[HTML]{DAE8FC}51.6 & \cellcolor[HTML]{DAE8FC}65.3 & \cellcolor[HTML]{DAE8FC}51.5 & 59.42 \\
ORE & 63.5 & 70.9 & 58.9 & 42.9 & 34.1 & 76.2 & 80.7 & 76.3 & 34.1 & 66.1 & \cellcolor[HTML]{DAE8FC}56.1 & \cellcolor[HTML]{DAE8FC}70.4 & \cellcolor[HTML]{DAE8FC}80.2 & \cellcolor[HTML]{DAE8FC}72.3 & \cellcolor[HTML]{DAE8FC}81.8 & \cellcolor[HTML]{DAE8FC}42.7 & \cellcolor[HTML]{DAE8FC}71.6 & \cellcolor[HTML]{DAE8FC}68.1 & \cellcolor[HTML]{DAE8FC}77 & \cellcolor[HTML]{DAE8FC}67.7 & \textbf{64.58} \\ \midrule
\midrule
{\color[HTML]{009901} \textbf{15 + 5 setting}} & aero & cycle & bird & boat & bottle & bus & car & cat & chair & cow & table & dog & horse & bike & person & plant & sheep & sofa & train & tv & mAP \\ \midrule
First 15 & 74.2 & 79.1 & 71.3 & 60.3 & 60 & 80.2 & 88.1 & 80.2 & 48.8 & 74.6 & 61 & 76 & 85.3 & 78.2 & 83.4 & \cellcolor[HTML]{DAE8FC}0 & \cellcolor[HTML]{DAE8FC}0 & \cellcolor[HTML]{DAE8FC}0 & \cellcolor[HTML]{DAE8FC}0 & \cellcolor[HTML]{DAE8FC}0 & 55.03 \\
New 5 & 3.7 & 0.5 & 6.3 & 4.6 & 0.9 & 0 & 8.8 & 3.9 & 0 & 0.4 & 0 & 0 & 16.4 & 0.7 & 0 & \cellcolor[HTML]{DAE8FC}41 & \cellcolor[HTML]{DAE8FC}55.7 & \cellcolor[HTML]{DAE8FC}49.2 & \cellcolor[HTML]{DAE8FC}59.1 & \cellcolor[HTML]{DAE8FC}67.8 & 15.95 \\ \midrule
ILOD \cite{shmelkov2017incremental} & 70.5 & 79.2 & 68.8 & 59.1 & 53.2 & 75.4 & 79.4 & 78.8 & 46.6 & 59.4 & 59 & 75.8 & 71.8 & 78.6 & 69.6 & \cellcolor[HTML]{DAE8FC}33.7 & \cellcolor[HTML]{DAE8FC}61.5 & \cellcolor[HTML]{DAE8FC}63.1 & \cellcolor[HTML]{DAE8FC}71.7 & \cellcolor[HTML]{DAE8FC}62.2 & 65.87 \\
ILOD + Faster R-CNN & 63.5 & 76.3 & 70.7 & 53.1 & 55.8 & 67.1 & 81.5 & 80.3 & 49.6 & 73.8 & 62.1 & 77.1 & 79.7 & 74.2 & 73.9 & \cellcolor[HTML]{DAE8FC}37.1 & \cellcolor[HTML]{DAE8FC}59.1 & \cellcolor[HTML]{DAE8FC}61.7 & \cellcolor[HTML]{DAE8FC}68.6 & \cellcolor[HTML]{DAE8FC}61.3 & 66.35 \\
Faster ILOD \cite{PENG2020109} & 66.5 & 78.1 & 71.8 & 54.6 & 61.4 & 68.4 & 82.6 & 82.7 & 52.1 & 74.3 & 63.1 & 78.6 & 80.5 & 78.4 & 80.4 & \cellcolor[HTML]{DAE8FC}36.7 & \cellcolor[HTML]{DAE8FC}61.7 & \cellcolor[HTML]{DAE8FC}59.3 & \cellcolor[HTML]{DAE8FC}67.9 & \cellcolor[HTML]{DAE8FC}59.1 & 67.94 \\ \midrule
ORE - (CC  + EBUI) & 65.1 & 74.6 & 57.9 & 39.5 & 36.7 & 75.1 & 80 & 73.3 & 37.1 & 69.8 & 48.8 & 69 & 77.5 & 72.8 & 76.5 & \cellcolor[HTML]{DAE8FC}34.4 & \cellcolor[HTML]{DAE8FC}62.6 & \cellcolor[HTML]{DAE8FC}56.5 & \cellcolor[HTML]{DAE8FC}80.3 & \cellcolor[HTML]{DAE8FC}65.7 & 62.66 \\
ORE & 75.4 & 81 & 67.1 & 51.9 & 55.7 & 77.2 & 85.6 & 81.7 & 46.1 & 76.2 & 55.4 & 76.7 & 86.2 & 78.5 & 82.1 & \cellcolor[HTML]{DAE8FC}32.8 & \cellcolor[HTML]{DAE8FC}63.6 & \cellcolor[HTML]{DAE8FC}54.7 & \cellcolor[HTML]{DAE8FC}77.7 & \cellcolor[HTML]{DAE8FC}64.6 & \textbf{68.51} \\ \midrule
\midrule
{\color[HTML]{009901} \textbf{19 + 1 setting}} & aero & cycle & bird & boat & bottle & bus & car & cat & chair & cow & table & dog & horse & bike & person & plant & sheep & sofa & train & tv & mAP \\ \midrule
First 19 & 77.8 & 81.7 & 69.3 & 51.6 & 55.3 & 74.5 & 86.3 & 80.2 & 49.3 & 82 & 63.6 & 76.8 & 80.9 & 77.5 & 82.4 & 42.9 & 73.9 & 70.4 & 70.4 & \cellcolor[HTML]{DAE8FC}0 & 67.34 \\
Last 1 & 0 & 0 & 0 & 0 & 0 & 0 & 0 & 0 & 0 & 0 & 0 & 0 & 0 & 0 & 0 & 0 & 0 & 0 & 0 & \cellcolor[HTML]{DAE8FC}64 & 3.2 \\ \midrule
ILOD \cite{shmelkov2017incremental} & 69.4 & 79.3 & 69.5 & 57.4 & 45.4 & 78.4 & 79.1 & 80.5 & 45.7 & 76.3 & 64.8 & 77.2 & 80.8 & 77.5 & 70.1 & 42.3 & 67.5 & 64.4 & 76.7 & \cellcolor[HTML]{DAE8FC}62.7 & 68.25 \\
ILOD + Faster R-CNN & 60.9 & 74.6 & 70.8 & 56 & 51.3 & 70.7 & 81.7 & 81.5 & 49.45 & 78.3 & 58.3 & 79.5 & 79.1 & 74.8 & 75.7 & 42.8 & 74.7 & 61.2 & 67.2 & \cellcolor[HTML]{DAE8FC}65.1 & 67.72 \\
Faster ILOD \cite{PENG2020109} & 64.2 & 74.7 & 73.2 & 55.5 & 53.7 & 70.8 & 82.9 & 82.6 & 51.6 & 79.7 & 58.7 & 78.8 & 81.8 & 75.3 & 77.4 & 43.1 & 73.8 & 61.7 & 69.8 & \cellcolor[HTML]{DAE8FC}61.1 & 68.56 \\ \midrule
ORE - (CC  + EBUI) & 60.7 & 78.6 & 61.8 & 45 & 43.2 & 75.1 & 82.5 & 75.5 & 42.4 & 75.1 & 56.7 & 72.9 & 80.8 & 75.4 & 77.7 & 37.8 & 72.3 & 64.5 & 70.7 & \cellcolor[HTML]{DAE8FC}49.9 & 64.93 \\
ORE & 67.3 & 76.8 & 60 & 48.4 & 58.8 & 81.1 & 86.5 & 75.8 & 41.5 & 79.6 & 54.6 & 72.8 & 85.9 & 81.7 & 82.4 & 44.8 & 75.8 & 68.2 & 75.7 & \cellcolor[HTML]{DAE8FC}60.1 & \textbf{68.89} \\ \bottomrule
\end{tabular}%
}\vspace{-0.5em}
\caption{
We compare \method against  state-of-the-art incremental Object Detectors on three different settings. $10$, $5$ and the last class from the Pascal VOC 2007 \cite{everingham2010pascal} dataset are introduced to a detector trained on $10$, $15$ and $19$ classes respectively (shown in \colorbox{ iODBlue}{blue} background). \method is able to perform favourably on all the settings with no methodological change. Kindly refer to Sec.~\ref{sec:incr_OD_results} for more details.}
\label{tab:iOD}
\vspace{-13pt}
\end{table*}

We find an interesting consequence of the ability of \method to distinctly model unknown objects: it performs favorably well on the incremental object detection (iOD) task against the state-of-the-art (Tab.~\ref{tab:iOD}). This is because, \method reduces the confusion of an unknown object being classified as a known object, which lets the detector incrementally learn the true foreground objects. We use the standard protocol \cite{shmelkov2017incremental,PENG2020109} used in the iOD domain to evaluate \method, where group of classes ($10$, $5$ and the last class) from Pascal VOC 2007 \cite{everingham2010pascal} are incrementally learned by a detector trained on the remaining set of classes. Remarkably, \method is used as it is, without any change to the methodology introduced in Sec.~\ref{sec:ore_methodology}. We ablate contrastive clustering (CC) and energy based unknown identification (EBUI) to find that it results in reduced performance than standard \method.

\section{Discussions and Analysis}\label{sec:discussions_and_analysis}

\customsubsection{Ablating \method Components:}\label{sec:ablation} To study the contribution of each of the components in \method, we design careful ablation experiments (Tab.~\ref{tab:ablation}). We consider the setting where Task 1 is introduced to the model. The auto-labelling methodology (referred to as ALU), combined with energy based unknown identification (EBUI) performs better together (row $5$) than using either of them separately (row $3$ and $4$). Adding contrastive clustering (CC) to this configuration, gives the best performance in handling unknown (row $7$), measured in terms of WI and A-OSE. There is no severe performance drop in known classes detection (mAP metric) as a side effect of unknown identification. In row $6$, we see that EBUI is a critical component whose absence increases WI and A-OSE scores. Thus, each component in \method has a critical role to play for unknown identification.

\begin{table}
\centering
\resizebox{0.48\textwidth}{!}{%
\begin{tabular}{@{}c|ccc|ccc@{}}
\toprule
Row ID & CC & ALU & EBUI & WI ($\downarrow$) & A-OSE ($\downarrow$) & mAP ($\uparrow$) \\ \midrule
1 &  & Oracle &  & 0.02004 & 7080 & 57.76 \\ \midrule
2 & $\times$ & $\times$ & $\times$ & 0.06991 & 13396 & {56.16} \\
3 & $\times$ & $\times$ & \checkmark & 0.05932 & 12822 & 56.21 \\
4 & $\times$ & \checkmark & $\times$ & 0.05542 & 12111 & 56.09 \\
5 & $\times$ & \checkmark & \checkmark & 0.04539 & 9011 & 55.95 \\
6 & \checkmark & \checkmark & $\times$ & 0.05614 & 12064 & \textbf{56.36} \\
7 & \checkmark & \checkmark & \checkmark & \textbf{0.02193} & \textbf{8234} & 56.34 \\ \bottomrule
\end{tabular}%
}\vspace{-0.5em}
\caption{We carefully ablate each of the constituent component of \method. CC, ALU and EBUI refers to `Contrastive Clustering', `Auto-labelling of Unknowns' and `Energy Based Unknown Identifier' respectively. Kindly refer to Sec.~\ref{sec:ablation} for more details.}
\label{tab:ablation}
\end{table}

\customsubsection{Sensitivity Analysis on Exemplar Memory Size:}
Our balanced finetuning strategy requires storing exemplar images with at least $N_{ex}$ instances per class. We vary $N_{ex}$ while learning Task 2 and report the results in Table \ref{tab:memory_size}. We find that balanced finetuning is very effective in improving the accuracy of previously known class, even with just having minimum $10$ instances per class. However, we find that increasing $N_{ex}$ to large values does-not help and at the same time adversely affect how unknowns are handled (evident from WI and A-OSE scores). Hence, by validation, we set $N_{ex}$ to $50$ in all our experiments, which is a sweet spot that balances performance on known and unknown classes.

\begin{table}
\centering
\resizebox{0.48\textwidth}{!}{%
\begin{tabular}{@{}c|cc|ccc@{}}
\toprule
$N_{ex}$ & WI & A-OSE & \multicolumn{3}{c}{mAP ($\uparrow$)} \\ \midrule
 & ($\downarrow$) & ($\downarrow$) & \begin{tabular}[c]{@{}c@{}}Previously  known\end{tabular} & \begin{tabular}[c]{@{}c@{}}Current known\end{tabular} & Both \\ \cmidrule(l){4-6} 
0 & 0.0406 & 9268 & 8.74 & 26.81 & 17.77 \\
10 & 0.0237 & 8211 & 46.78 & 24.32 & 35.55 \\
20 & 0.0202 & 8092 & 48.83 & 25.42 & 37.13 \\
\cellcolor[HTML]{E6FFE1}50 & \cellcolor[HTML]{E6FFE1}0.0154 & \cellcolor[HTML]{E6FFE1}7772 & \cellcolor[HTML]{E6FFE1}52.37 & \cellcolor[HTML]{E6FFE1}25.58 & \cellcolor[HTML]{E6FFE1}38.98 \\
100 & 0.0410 & 11065 & 52.29 & 26.21 & 39.24 \\
200 & 0.0385 & 10474 & 53.41 & 26.35 & 39.88 \\
400 & 0.0396 & 11461 & 53.18 & 26.09 & 39.64 \\ \bottomrule
\end{tabular}%
}\vspace{-0.5em}
\caption{The table shows sensitivity analysis. Increasing $N_{ex}$ by a large value hurts performance on unknown, while a small set of images are essential to mitigate forgetting (best row in \colorbox{ green!10}{green}).}
\label{tab:memory_size}
\end{table}

\customsubsection{Comparison with an \OS Detector:}
The mAP values of the detector when it is evaluated on closed set data (trained and tested on Pascal VOC 2007) and \os data (test set contains equal number of unknown images from MS-COCO) helps to measure how the detector handles unknown instances. Ideally, there should not be a performance drop. We compare \method against the recent \os detector proposed by Miller \etal \cite{miller2018dropout}. 
We find from Tab.~\ref{tab:comparison_with_OS} that drop in performance of \method is much lower than \cite{miller2018dropout} owing to the effective modelling of the unknown instances.

\begin{table}
\centering\setlength{\tabcolsep}{8pt}
\resizebox{0.48\textwidth}{!}{%
\begin{tabular}{@{}l|cc@{}}
\toprule
Evaluated on $\rightarrow$ & VOC 2007 & VOC 2007 + COCO (WR1) \\ \midrule
Standard Faster R-CNN & 81.86 & 77.09 \\
Dropout Sampling \cite{miller2018dropout} & 78.15 & 71.07 \\
ORE & \textbf{81.31} & \textbf{78.16} \\ \bottomrule
\end{tabular}%
}\vspace{-0.5em}
\caption{Performance comparison with an \OS object detector. \method is able to reduce the fall in mAP values considerably.}
\label{tab:comparison_with_OS}\vspace{-0.2em}
\end{table}

\begin{figure}
\includegraphics[width=\columnwidth]{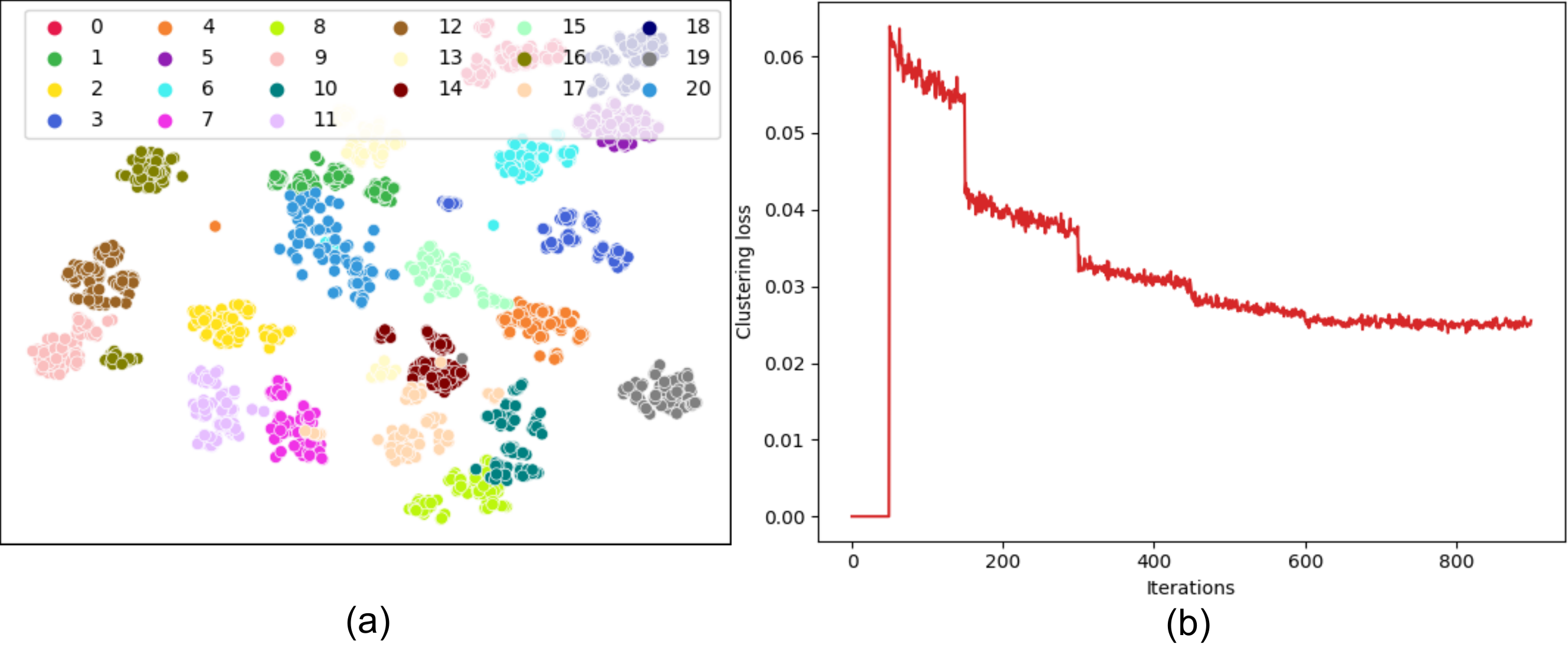}
\vspace{-25pt}
\caption{(a) Distinct clusters in the latent space.  (b) Our contrastive loss which ensures such a clustering steadily converges.}
\label{fig:tsne_loss}\vspace{-0.3em}
\end{figure}

\customsubsection{Clustering loss and t-SNE \cite{maaten2008visualizing} visualization:} We visualise the quality of clusters that are formed while training with the contrastive clustering loss (Eqn.~\ref{eqn:clustering_loss}) for Task 1. We see nicely formed clusters in Fig.~\ref{fig:tsne_loss} (a). Each number in the legend correspond to the $20$ classes introduced in Task 1. Label $20$ denotes unknown class. Importantly, we see that the unknown instances also gets clustered, which reinforces the quality of the auto-labelled unknowns used in contrastive clustering. In Fig.~\ref{fig:tsne_loss} (b), we plot the contrastive clustering loss against training iterations, where we see a gradual decrease, indicative of good convergence. 

\begin{figure}
\includegraphics[width=\columnwidth]{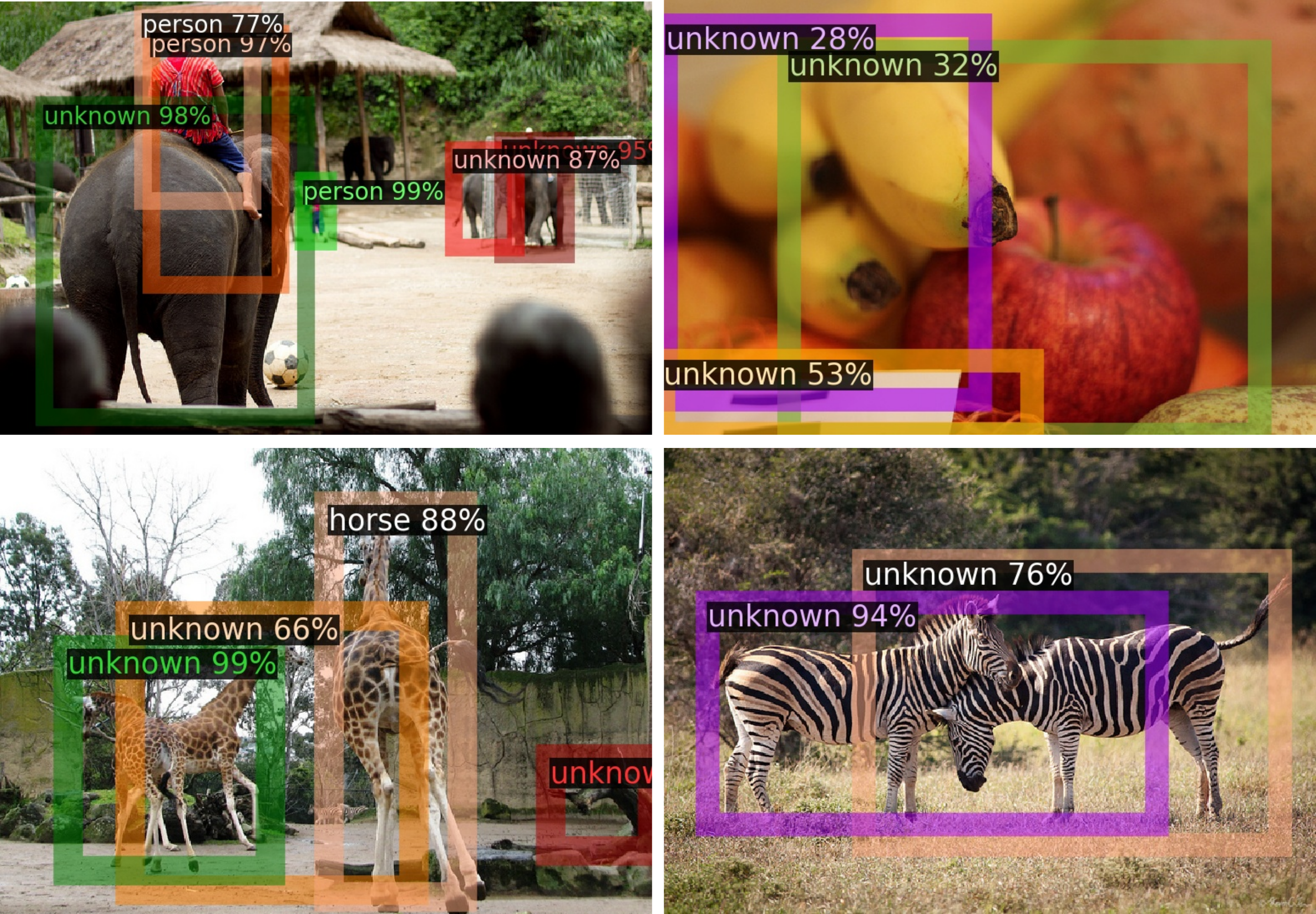}
\vspace{-18pt}
\caption{Predictions from \method after being trained on Task 1. `\cls{elephant}', `\cls{apple}', `\cls{banana}', `\cls{zebra}' and `\cls{giraffe}' have not been introduced to the model, and hence are successfully classified as `\cls{unknown}'. The approach misclassifies one of the `\cls{giraffe}' as a `\cls{horse}', showing the limitation of \method. }
\label{fig:qual_results}
\end{figure}

\section{Conclusion}\label{sec:conclusion}\vspace{-0.5em}
The vibrant object detection community has pushed the performance benchmarks on standard datasets by a large margin. 
The closed-set nature of these datasets and evaluation protocols, hampers further progress. We introduce \OWOD, where the object detector is able to label an unknown object as unknown and gradually learn the unknown as the model gets exposed to new labels. Our key novelties include an energy-based classifier for unknown detection and a contrastive clustering approach for open world learning.  We hope that our work will kindle further research along this important and open direction.






\vspace{-10pt}
\section*{Acknowledgements}\label{sec:Acknowledgements}\vspace{-0.5em}
\noindent We thank TCS for supporting KJJ through its PhD fellowship; MBZUAI for a start-up grant; VR starting grant (2016-05543) and DST, Govt of India, for partly supporting this work through IMPRINT program (IMP/2019/000250). We thank our anonymous reviewers for their valuable feedback.

{\small
\bibliographystyle{ieee_fullname}
\bibliography{egbib}
}


\clearpage
\appendix
\section*{\centering Supplementary Material}

In this supplementary material, we provide additional details which we could not include in the main paper due to space constraints, including experimental analysis, implementation details, discussion and results that help us develop further insights to the proposed \OWOD approach. We discuss:
\begin{itemize}
\setlength\itemsep{-0.15em}
    \item Sensitivity analysis on queue size of Feature Store, the momentum parameter $\eta$,  margin in clustering loss $\Delta$ and  temperature parameter in energy computation.
    \item Additional details on contrastive clustering
    \item More specific implementation details.
    \item Discussion regarding failure cases.
    \item Related works in incremental object detection.
    \item Some qualitative results of \method.
\end{itemize}

\section{Varying the Queue Size of $\mathcal{F}_{Store}$}
In Sec.~4.1, we explain how class specific queues $\bm{q}_i$ are used to store the feature vectors, which are used to compute the class prototypes. A hyper-parameter $\nQ$ controls the size of each $\bm{q}_i$. Here we vary $\nQ$, while learning Task 1, and report the results in Tab.~\ref{tab:varying_q}. We observe relatively similar performance, across experiments with different $\nQ$ values. This can be attributed to the fact that after a prototype is defined, it gets periodically updated with newly observed features, thus effectively evolving itself. Hence, the actual number of features used to compute those prototypes ($\mathcal{P} \text{ and } \mathcal{P}_{new}$) is not very significant. We use $\nQ = 20$ for all the experiments.

\begin{table}[h]
\centering
\resizebox{0.25\textwidth}{!}{%
\begin{tabular}{@{}c|ccc@{}}
\toprule
$\nQ$ & WI ($\downarrow$) & A-OSE ($\downarrow$) & mAP ($\uparrow$) \\ \midrule
5 & 0.02402 & 8123 & 56.01 \\
10 & 0.02523 & 8126 & 56.02 \\
20 & 0.02193 & 8234 & 56.34 \\
30 & 0.02688 & 8487 & 55.78 \\
50 & 0.02623 & 8578 & 56.22 \\ \bottomrule
\end{tabular}%
}
\caption{We find that varying the number of features that are used to compute the class prototype does not have a huge impact on the performance. }
\label{tab:varying_q}
\end{table}

\section{Sensitivity Analysis on $\eta$}
The momentum parameter $\eta$ controls how rapidly the class prototypes are updated, as elaborated in Algorithm 1. Larger values of $\eta$ imply smaller effect of the newly computed prototypes on the current class prototypes. We find from Tab.~\ref{tab:momentum} that performance improves when prototypes are updated slowly (larger values of $\eta$). This result is intuitive, as slowly changing the cluster centers helps stabilize contrastive learning.  

\begin{table}[h]
\centering
\resizebox{0.25\textwidth}{!}{%
\begin{tabular}{@{}c|ccc@{}}
\toprule
$\eta$ & WI ($\downarrow$) & A-OSE ($\downarrow$) & mAP ($\uparrow$) \\ \midrule
0.4 & 0.05926 & 9476 & 55.96 \\
0.6 & 0.04977 & 9095 & 55.56 \\
0.8 & 0.02945 & 8375 & 55.73 \\
0.9 & 0.02193 & 8234 & 56.34 \\ \bottomrule
\end{tabular}%
}
\caption{We see that higher values of $\eta$ gives better performance, implying that gradual evolution of class prototypes improves contrastive clustering.}
\label{tab:momentum}
\end{table}

\section{Varying the Margin ($\Delta$) in $\mathcal{L}_{cont}$}
The margin parameter $\Delta$ in the contrastive clustering loss $\mathcal{L}_{cont}$ (Eqn.~1) defines the minimum distance that an input feature vector should keep from dissimilar class prototypes in the latent space. As we see in Tab.~\ref{tab:delta}, increasing the margin while learning the first task, increases the performance on the known classes and how unknown classes are handled. This would imply that larger separation in the latent space is beneficial for \method.

\begin{table}[h]
\centering
\resizebox{0.25\textwidth}{!}{%
\begin{tabular}{@{}c|ccc@{}}
\toprule
$\Delta$ & WI ($\downarrow$) & A-OSE ($\downarrow$) & mAP ($\uparrow$) \\ \midrule
5 & 0.04094 & 9300 & 55.73 \\
10 & 0.02193 & 8234 & 56.34 \\
15 & 0.01049 & 8088 & 56.65 \\ \bottomrule
\end{tabular}%
}
\caption{Increasing the margin $\Delta$, improves the performance on known and unknown classes, concurring with our assumption that separation in the latent space is beneficial for \method.}
\label{tab:delta}
\end{table}

 

\section{Varying the Temperature ($T$) in Eqn.~\ref{eqn:energy}}

We fixed the temperature parameter ($T$) in Eqn.~\ref{eqn:energy} to 1 in all the experiments. Softening the energies a bit more to $T = 2$, gives slight improvement in unknown detection, however increasing it further hurts as evident from Tab.~\ref{tab:temperature}.
\begin{table}[h]
\centering
\resizebox{0.25\textwidth}{!}{%
\begin{tabular}{@{}c|ccc@{}}
\toprule
T & WI($\downarrow$) & A-OSE($\downarrow$)  & mAP($\uparrow$) \\ \midrule
1 & 0.0219 & 8234 & {56.34} \\
2 & {0.0214} & {8057} & 55.68 \\
3 & 0.0411 & 11266 & 55.51 \\
5 & 0.0836 & 12063 & 56.25 \\
10 & 0.0835 & 12064 & 56.31 \\ \bottomrule
\end{tabular}%
}
\caption{There is a nice ballpark for temperature parameter between $T=1$ and $T=2$, which gives the optimal performance.}
\label{tab:temperature}
\end{table}

\section{More Details on Contrastive Clustering}
The motivation for using contrastive clustering to ensure separation in the latent space 
is two-fold: 1) it enables the model to cluster unknowns separately from known instances, thus boosting unknown identification; 2) it ensures instances of each class are well-separated from other classes, alleviating the forgetting issue.
\begin{figure}[h]
\includegraphics[width=1\columnwidth]{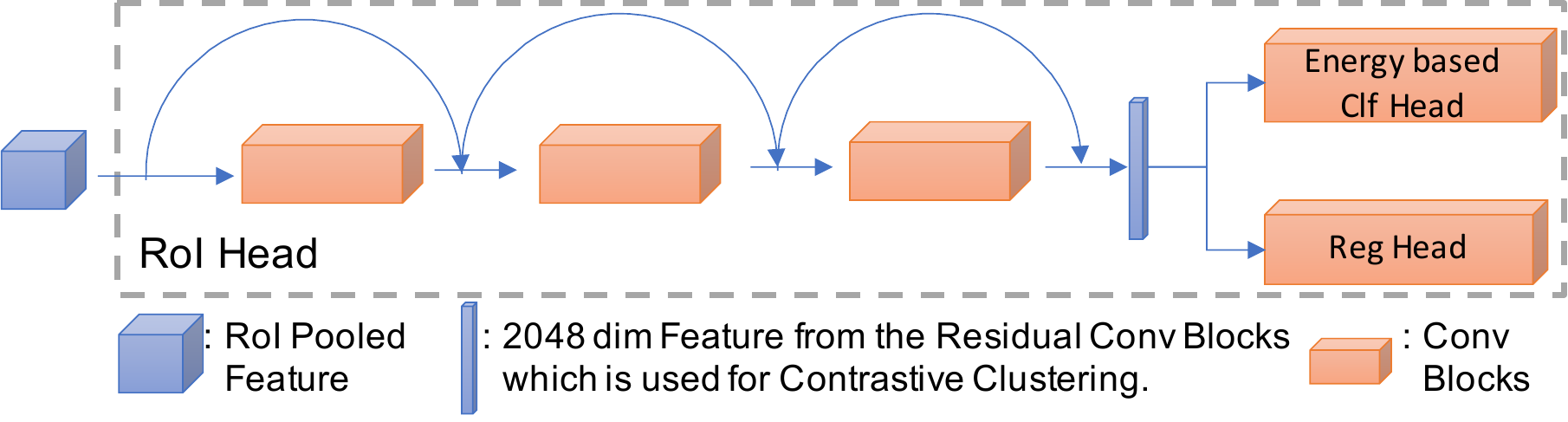}
\vspace{-10pt}
\caption{RoI head architecture, showing 2048-dim feature vector used for contrastive clustering.}
\vspace{-5pt}
\label{fig:contrastive_clustering}
\end{figure} 

The 2048-dim feature vector that comes out from residual blocks of RoI head (Fig~\ref{fig:contrastive_clustering}) is contrastively clustered. The contrastive loss is added to the Faster R-CNN loss and the entire network is trained end-to-end. Thus all parts of the network before and including the residual block in the RoI head in the Faster R-CNN pipeline will get updated with the gradients from the contrastive clustering loss. 

\section{Further Implementation Details} 
We complete the discussion related to the implementation details, that we had in Sec.~5.2 here. 
We ran our experiments on a server with $8$ Nvidia V100 GPUs 
with an effective batch size of $8$. We use SGD with a learning rate of $0.01$. Each task is learned for $8$ epochs ($\sim 50$k iterations). The queue size of the feature store is set to $20$. We initiate clustering after $1$k iterations and update the cluster prototypes after each $3$k iterations with a momentum parameter of $0.99$. Euclidean distance is used as the distance function $\mathcal{D}$ in Eqn.~1. The margin ($\Delta$)  is set as $10$. For auto-labelling the unknowns in the RPN, we pick the top-1 background proposal, sorted by its objectness score. The temperature parameter in the energy based classification head is set to $1$. The code is implemented in PyTorch \cite{NEURIPS2019_9015} using Detectron 2 \cite{wu2019detectron2}. Reliability  library \cite{matthew_reid_2020_4255226} was used for modelling the energy distributions.
We release all our codes publicly for foster reproducible research: \small\url{https://github.com/JosephKJ/OWOD}.

\section{Related Work on Incremental Object Detection} 
The class-incremental object detection (iOD) setting considers classes to be observed incrementally over time and that the learner must adapt without retraining on old classes from scratch. The prevalent approaches \cite{shmelkov2017incremental,li2019rilod,hao2019end,chen2019new} use knowledge distillation \cite{hinton2015distilling}  as a regularization measure to avoid forgetting old class information while training on new classes. Specifically, Shmelkov \etal~\cite{shmelkov2017incremental} repurpose Fast R-CNN for incremental learning by distilling classification and regression outputs from a previous stage model. Beside distilling model outputs, Chen \etal~\cite{chen2019new} and Li \etal~\cite{li2019rilod} also distilled the intermediate network features. Hao \etal~\cite{hao2019end} builds on Faster R-CNN and uses a student-teacher framework for RPN adaptation. 
Acharya \etal \cite{acharya2020rodeo} proposes a replay mechanism for online detection. 
Recently, Peng \etal~\cite{PENG2020109} introduces an adaptive distillation technique into Faster R-CNN. Their methodology is the current state-of-the-art in iOD.
These methods cannot however work in an \OW environment, which is the focus of this work, and are unable to identify unknown objects. 

%

\section{Time and Storage Expense:}
The training and inference of \method takes an additional $0.1349$ sec/iter and $0.009$ sec/iter than standard Faster R-CNN.  The storage expense for maintaining $F_{Store}$ is negligible, and the exemplar memory (for $N_{ex} = 50$) takes approximately $34$ MB.

\section{Using Softmax based Unknown Identifier} We modified the unknown identification criteria to \textit{max(softmax(logits)) $<$} t. For t = $\{0.3, 0.5, 0.7\}$: A-OSE, WI and mAP (mean and std-dev) are $11815 \pm 352.13$, $0.0436 \pm 0.009$ and $55.22 \pm 0.02$. This is inferior to \method.

\section{Qualitative Results}
\begin{figure*}[h]
\vspace{-30pt}
\centering
\subfloat[]{\includegraphics[width=0.49\linewidth]{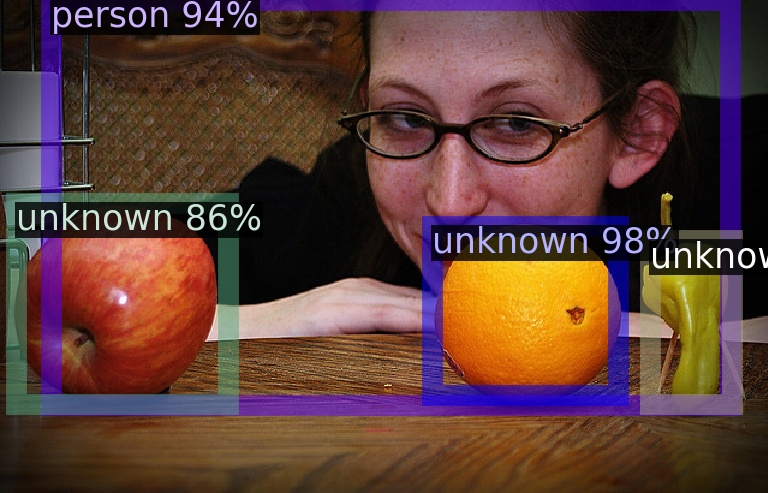}}
\hspace{3pt}\subfloat[]{\includegraphics[width=0.49\linewidth]{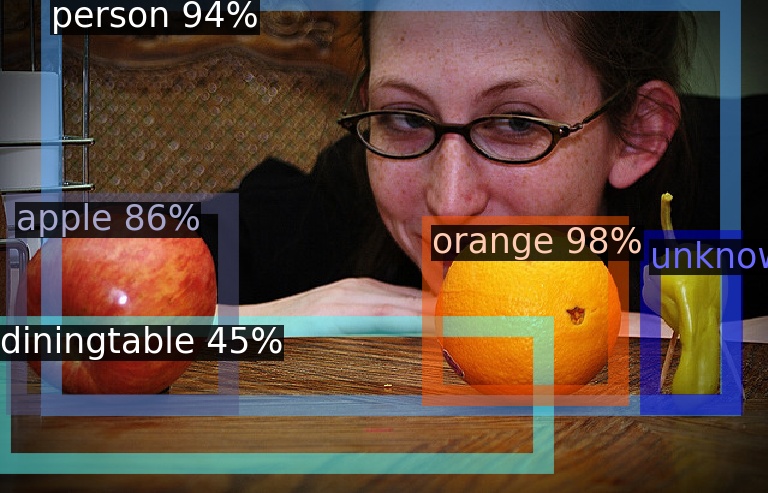}}
\vspace{-10pt}
\caption{The sub-figure (a) is the result produced by \method after learning Task 2. As Task 3 classes like \cls{apple} and \cls{orange} has not been introduced, \method identifies it and correctly labels them as \cls{unknown}. After learning Task 3, these instances are labelled correctly in sub-figure (b). An unidentified class instance still remains, and \method successfully detects it as \cls{unknown}.}
\label{fig:qual_1}
\end{figure*}

\begin{figure*}[h]
\vspace{-15pt}
\centering
\subfloat[]{\includegraphics[width=0.49\linewidth]{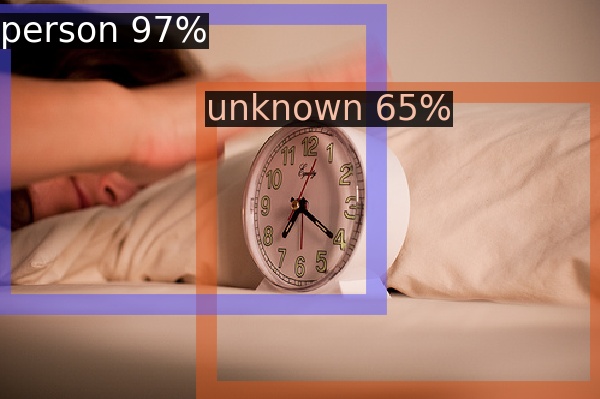}}
\hspace{3pt}\subfloat[]{\includegraphics[width=0.49\linewidth]{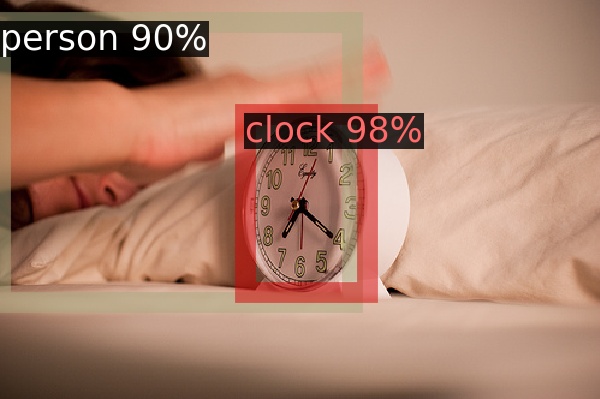}}
\vspace{-10pt}
\caption{The \cls{clock} class is eventually learned as part of Task 4 (in sub-figure (b)), after being initially identified as \cls{unknown} (in sub-figure (a)). \method  exhibits the true characteristics of an \OW detector, where it is able to incrementally learn an identified unknown.}
\label{fig:qual_3}
\end{figure*}

\begin{figure*}[h]
\vspace{-15pt}
\centering
\subfloat[]{\includegraphics[width=0.49\linewidth]{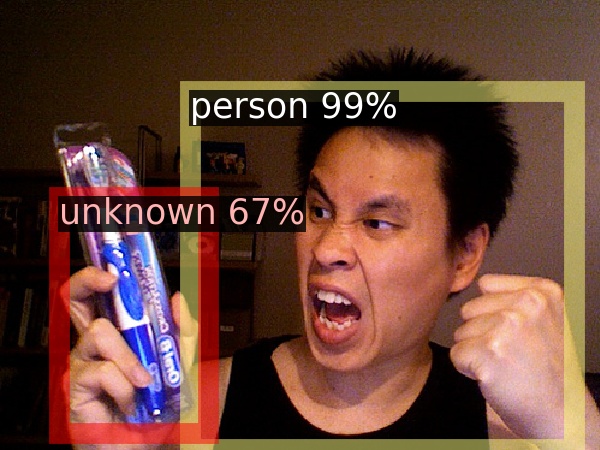}}
\hspace{3pt}\subfloat[]{\includegraphics[width=0.49\linewidth]{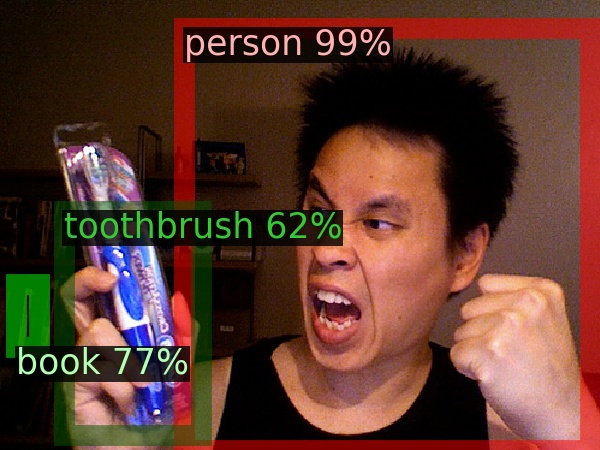}}
\vspace{-10pt}
\caption{\cls{toothbrush} and \cls{book} are indoor objects introduced as part of Task 4. The detector trained till Task 3, identifies \cls{toothbrush} as an unknown objects in sub-figure (a) and eventually learn it as part of Task 4, without forgetting how to identify \cls{person} in sub-figure (b).}
\label{fig:qual_2}
\end{figure*}

We show qualitative results of \method in Fig.~\ref{fig:qual_1} through Fig.~\ref{fig:qual_7}. We see that \method is able to identify a variety of unknown instances and incrementally learn them, using the proposed contrastive clustering and energy-based unknown identification methodology. Sub-figure (a) in all these images shows the identified unknown instances along with the the other instances known to the detector. The corresponding sub-figure (b), shows the detections from the same detector after the new classes are incrementally added.

\section{Discussion Regarding Failure Cases}
\begin{figure*}[h]
\centering
\subfloat[]{\includegraphics[width=0.49\linewidth]{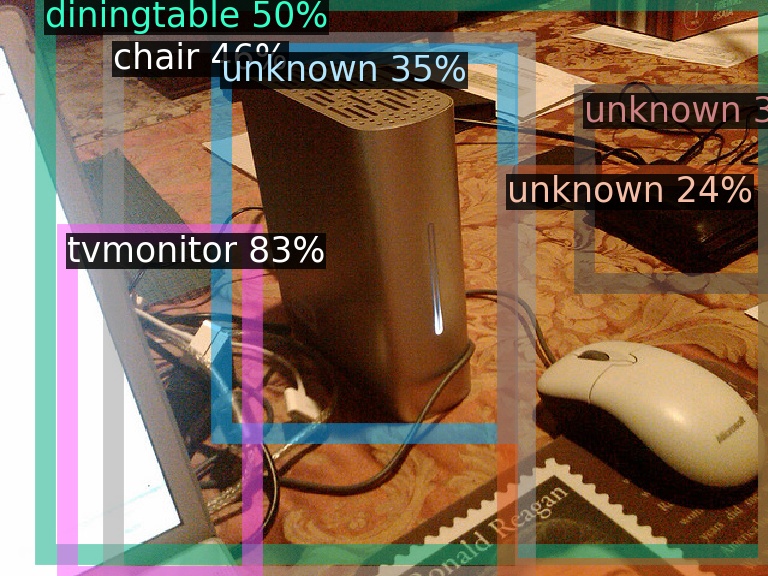}}
\hspace{3pt}\subfloat[]{\includegraphics[width=0.49\linewidth]{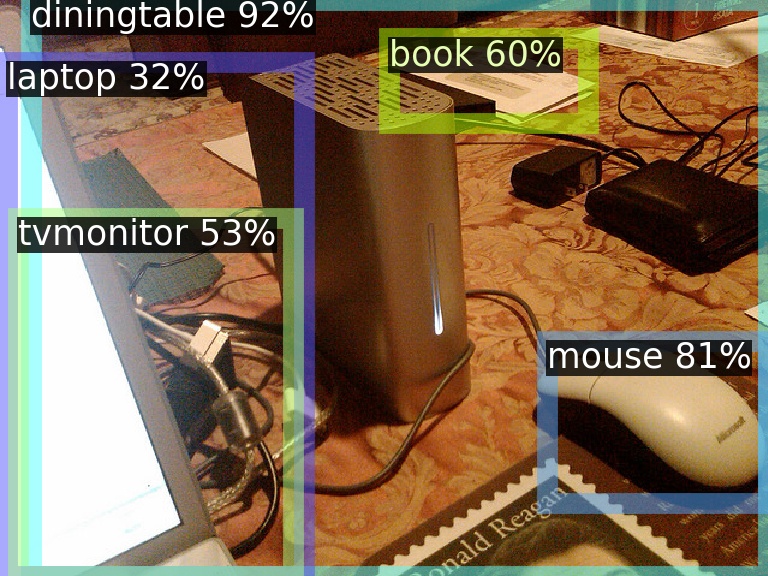}}
\vspace{-5pt}
\caption{Several items next to a \cls{laptop} on top of a table are identified as  \cls{unknown}, after learning Task 1. \cls{laptop}, \cls{book} and \cls{mouse} are introduced as part of Task 4, and hence are detected afterwords. \cls{external-storage} and \cls{walkman} (both are never introduced) were identified as \cls{unknown} initially, but has not been detected after learning  Task 4, and is one of the failure cases of \method.}
\label{fig:qual_5}
\end{figure*}

\begin{figure*}[h]
\centering
\subfloat[]{\includegraphics[width=0.49\linewidth]{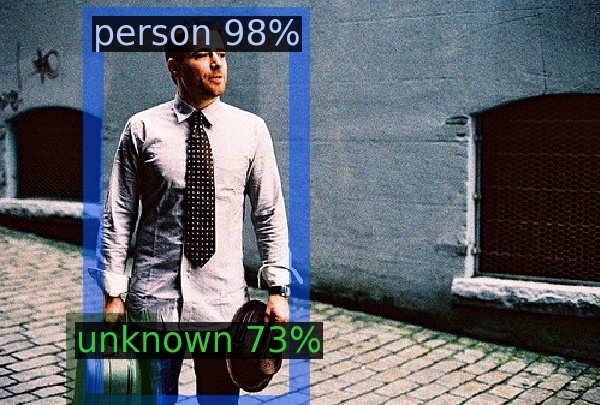}}
\hspace{3pt}\subfloat[]{\includegraphics[width=0.49\linewidth]{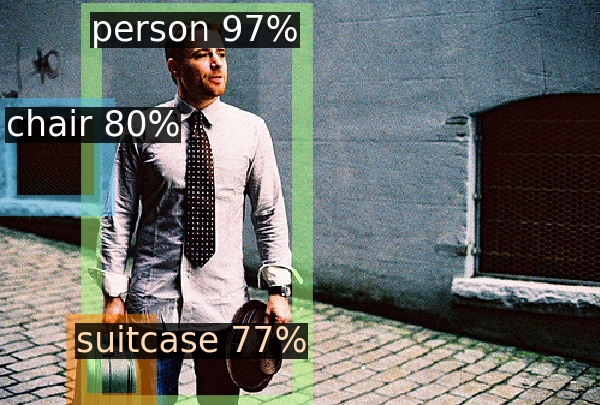}}
\vspace{-5pt}
\caption{\cls{suitcase} which was identified as \cls{unknown} is eventually learned in Task 2, along with a false positive detection of \cls{chair}.}
\label{fig:qual_6}
\end{figure*}

\begin{figure*}[h]
\centering
\subfloat[]{\includegraphics[width=0.49\linewidth]{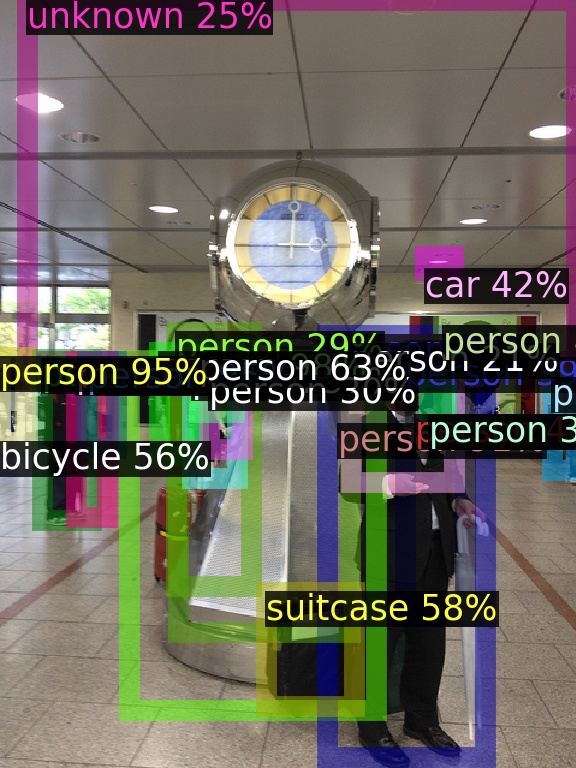}}
\hspace{3pt}\subfloat[]{\includegraphics[width=0.49\linewidth]{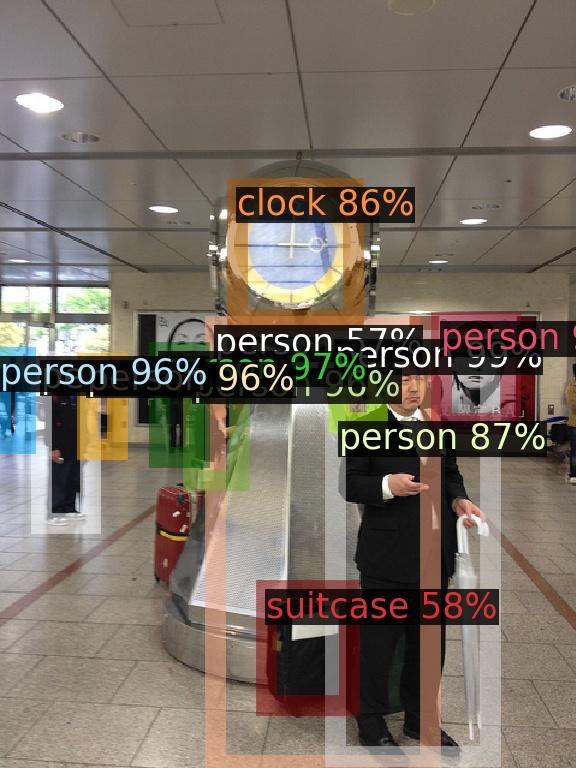}}
\vspace{-5pt}
\caption{In this highly cluttered scene, the \cls{unknown} instance \cls{clock} is identified, but is not localised well, after learning Task 2. After learning Task 4, \method detects \cls{clock}, along with reducing false positive detections of \cls{car} and \cls{bicycle}. The red \cls{suitcase} is not labelled after learning either of the tasks, and hence is a failure case.}
\label{fig:qual_7}
\end{figure*}

Occlusions and crowding of objects are cases where our method tends to get confused (\cls{external-storage}, \cls{walkman} and \cls{bag} not detected as \cls{unknown} in Figs.~\ref{fig:qual_5},~\ref{fig:qual_7}). Difficult viewpoints (such as backside) also lead to some misclassifications (\cls{giraffe}$\rightarrow$\cls{horse} in Figs.~\ref{fig:qual_results},~\ref{fig:qual_6}).
We have also noticed that detecting small \emph{unknown} objects co-occurring with larger known objects is hard. 
As \method is the first effort in this direction, we hope these identified shortcomings would be basis of further research. 

\begin{figure}[h]
\vspace{-5pt}
\centering
\subfloat[]{\includegraphics[width=0.9\linewidth]{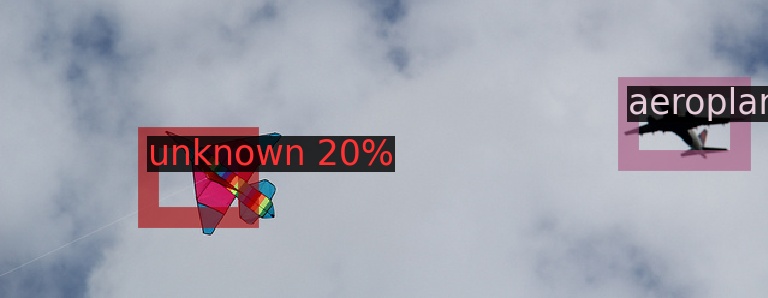}}
\hspace{3pt}\subfloat[]{\includegraphics[width=0.9\linewidth]{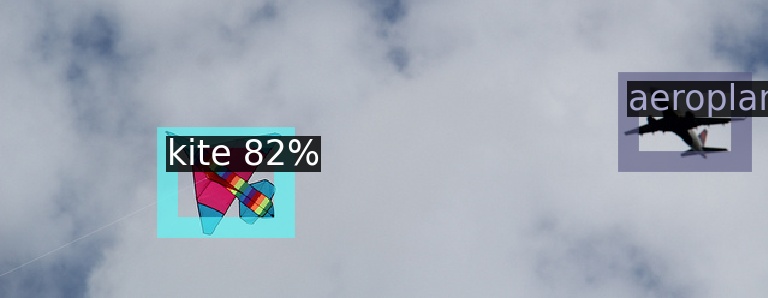}}
\caption{\method trained on just Task 1, successfully localises a \cls{kite} as an unknown in sub-figure (a), while after learning about \cls{kite} in Task 3, it incrementally learns to detect both \cls{kite} and \cls{aeroplane} in sub-figure (b).}
\label{fig:qual_4}
\end{figure}


\end{document}